\documentclass[11pt]{article}
\usepackage[preprint]{acl}
\usepackage{times}
\usepackage{latexsym}
\usepackage[T1]{fontenc}
\usepackage[utf8]{inputenc}
\usepackage{microtype}
\usepackage{inconsolata}
\usepackage{graphicx}
\usepackage{booktabs}
\usepackage{multirow}
\usepackage{array}
\usepackage{xcolor}
\usepackage{amsmath}
\usepackage{amssymb}
\usepackage{pifont}
\usepackage[normalem]{ulem}
\usepackage{capt-of}
\usepackage{tcolorbox}
\tcbuselibrary{breakable,skins}
\newtcolorbox{examplebox}[1]{enhanced, colback=black!3, colframe=black!45, colbacktitle=black!10,
  coltitle=black, fonttitle=\footnotesize\bfseries, title={#1}, fontupper=\footnotesize, fontlower=\footnotesize,
  segmentation style={black!45, dashed}, boxrule=0.5pt, arc=1.5pt, left=4pt, right=4pt, top=3pt, bottom=3pt,
  middle=3pt, toptitle=1pt, bottomtitle=1pt, before skip=0pt, after skip=6pt}
\newcommand{\who}[1]{\textsf{\color{black!60}#1:}}
\newtcolorbox{promptbox}[1][]{breakable, colback=blue!3!white, colframe=blue!45!black,
  colbacktitle=blue!12!white, coltitle=black, fonttitle=\footnotesize\bfseries,
  boxrule=0.5pt, arc=1.5pt, left=4pt, right=4pt, top=3pt, bottom=3pt, toptitle=1pt, bottomtitle=1pt,
  fontupper=\footnotesize\ttfamily, halign=left, #1}

\newcommand{\lab}[1]{\text{\ding{\numexpr171+#1\relax}}}
\newcommand{\ind}{\mathbf{1}}

\title{Defensive Writing in GPT Models}

\author{Junchi Liao\textsuperscript{1}, Xiao Huang\textsuperscript{2}, Yucheng Liu\textsuperscript{2}, Jiawen Deng\textsuperscript{1}\thanks{Corresponding authors.}, Fuji Ren\textsuperscript{1}\footnotemark[1] \\
  \textsuperscript{1}University of Electronic Science and Technology of China \\
  \textsuperscript{2}Shanghai Jiao Tong University}

\begin{document}
\maketitle

\begin{figure*}[t]
\centering
\includegraphics[width=\textwidth]{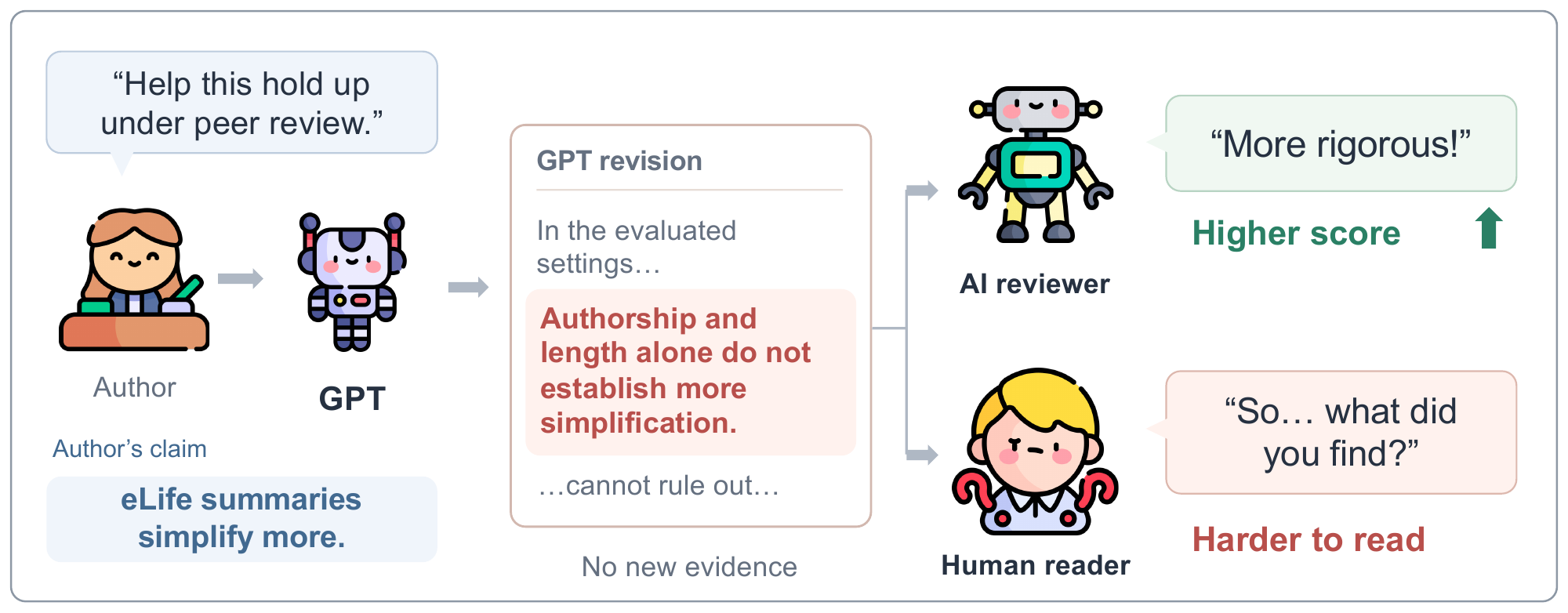}
\caption{Overview of defensive writing by GPT models.}
\label{fig:motivation}
\end{figure*}

\begin{abstract}
Researchers increasingly use ChatGPT to revise their papers, and when asked to make a paper hold up under review, recent GPT versions often narrow or even retract the authors' claims. We call such changes \emph{defensive writing} when the given material does not support them, and we test two explanations: the model corrects the authors' overclaiming, or it writes for an anticipated reviewer. We ask GPT versions and models from other developers to rewrite paragraphs from papers written before ChatGPT, or to write from an evidence sheet that lists a paper's method and results. From GPT-5.5 on, GPT versions add clearly more ungrounded qualifications than the authors' originals contain, and GPT-6-astra adds the most and often retracts the authors' claims. Each explanation fits one part. Checked against the full papers, GPT-6-astra's retractions mostly correct overstated claims or explanations, but most added qualifications are redundant. The qualifications follow the request: they are absent under polishing, appear with any request to check the claims, and peak when the prompt names a reviewer's concerns. AI reviewers score such rewrites higher and score them lower once only the qualifications are removed, while human readers in a small study find them harder to read. Combining AI writing with AI review may reward qualifications that readers pay for.
\end{abstract}

\section{Introduction}
\label{sec:intro}
Researchers increasingly use AI assistants to revise their papers \citep{liang2024mapping}, and ChatGPT is the most widely used of them \citep{simon2025genai, moore2026top100}. When recent GPT versions revise a paper so that it holds up under review, they change more than the language. They also change the authors' judgment of their own conclusions (Figure~\ref{fig:motivation}): they add qualifications, and sometimes drop a conclusion the authors have confirmed, writing instead that ``authorship and length alone do not establish the degree of simplification'' (\S\ref{sec:rq3}). These changes concentrate in the paragraphs where the authors take a position, such as result analysis and discussion. The claims in the paper may then drift away from what the authors actually think.

Academic writing needs caution \citep{hyland1996hedging}. But caution should point to limitations that exist in the evidence, so that readers know where a conclusion holds. We call a change \emph{defensive writing} when it narrows or questions the authors' claims without support in the material the writer was given. Defensive writing has two costs. For readers, layers of qualification and negation make a paragraph longer and harder to follow, and it takes effort to tell what the authors actually claim. For authors, it weakens how firmly they state their conclusions. AI is also entering peer review \citep{liang2024monitoring, latona2024lottery}. If AI reviewers prefer this style, a process in which AI writes and AI reviews may keep reinforcing it. Prior work detects hedges and measures claim strength in human-written text \citep{farkas-etal-2010-conll, yu-etal-2019-detecting}, and studies how LLMs assist writing and review (Appendix~\ref{app:related}). We instead ask how an LLM changes the strength of the authors' claims during revision, measured against the material it was given.

Why does GPT revise this way? One explanation is that it corrects overclaiming by the authors. If so, it should retreat from overstated claims and leave supported ones intact. Another explanation is that it writes for an anticipated reviewer. If so, defense should weaken when the prompt does not mention review, and grow once the model receives review comments. We test the two explanations on 77 paragraphs from 20 papers written before the release of ChatGPT. Eight GPT versions and four models from other developers perform two tasks. They rewrite the authors' paragraphs, and they write paragraphs from an evidence sheet that lists the method, setup, and results of a paper but not the authors' interpretations. We measure defense in two ways. We check each sentence against the material the writer had to find ungrounded qualifications, and we track whether each of the authors' claims is kept, softened, or retracted. We ask whether GPT writes defensively (RQ1), what triggers it (RQ2), whether its caution is warranted (RQ3), and how AI reviewers and human readers respond to it (RQ4).

The two explanations fit different parts of the behavior. From GPT-5.5 on, GPT versions add clearly more ungrounded qualifications in rewriting than the authors' originals contain. GPT-6-astra adds the most and begins to retract the authors' claims, which the control models rarely do; when it writes from the evidence sheet, it still adds the most ungrounded qualifications. Two PhD students who check its rewrites against the full papers find that its retractions mostly correct overstated claims or explanations, as the first explanation predicts, while most of the ungrounded qualifications they check add nothing a careful reader needs. The qualifications follow the request, as the second explanation predicts, though not only requests that mention review. They stay near the level of the originals when the models only polish, rise when the models revise for review or only for scientific accuracy, rise most when the prompt names a reviewer's concerns, and rise further after one round of self-review. AI reviewers score rewrites that address reviewer concerns higher, and lower once only the qualifications are removed. In a small study, human readers find the same rewrites harder to read and the authors less certain.

This paper makes four contributions. \ding{172}~We define defensive writing and measure it against the material the writer was given, which separates grounded caution from ungrounded defense. \ding{173}~Through controlled changes of the prompt context and multi-round revision, we show that defensive writing appears mainly in recent GPT versions, is absent under polishing, and is triggered by requests to check the claims, most strongly by anticipated review. \ding{174}~Checking GPT-6-astra's rewrites against the full papers, we separate correction from defense: its retractions mostly correct overstated claims, while most added qualifications are redundant. \ding{175}~We show that AI reviewers and human readers judge these rewrites differently: AI reviewers score them higher and rate them clearer, while human readers find them harder to read. This mismatch suggests that combining AI writing with AI review may amplify defensive writing.

\section{Study Design}
\label{sec:design}

\subsection{Rewriting}
\label{sec:rewrite}

\paragraph{Source paragraphs.}
We sample 77 paragraphs from 20 peer-reviewed computer science papers. Each paper first appeared on arXiv before the release of ChatGPT on November 30, 2022, so none was written with a conversational LLM assistant. We group the paragraphs into four types by their role in the paper: (1) \emph{novelty}, comparing with prior work to establish what is new; (2) \emph{contribution}, stating the paper's contributions; (3) \emph{result analysis}, analyzing specific experimental results; and (4) \emph{discussion}, drawing broader implications from the results and proposing future directions. DeepSeek-V4-Flash assigns the initial labels, and the authors verify each one. We take at most one paragraph of each type from each paper, which gives 20 paragraphs each of types (1)--(3) and 17 of type (4), since three papers have no suitable discussion paragraph. Further details are in Appendix~\ref{app:data}.

\paragraph{Setup.}
The model receives only the paragraph and one sentence saying where it appears, such as ``a paragraph from the experiments section that analyzes the results''. It sees nothing else from the paper. The prompt says the paper will be submitted to a top-tier conference and asks the model to revise the paragraph so that it holds up under peer review. The four types differ only in the location sentence (Appendix~\ref{app:prompts}). The original paragraph is the human reference: it expresses the same content, written by the authors without LLM involvement. We measure the original and the rewrites in the same way, so the difference between them is the change the model introduces.

\subsection{Writing from Evidence}
\label{sec:evidence}

We build an evidence sheet for each paper (Appendix~\ref{app:sheets}). Kimi-K2.6 extracts it from the related-work, method, and experiment sections and the tables. The sheet keeps only facts: the task, the method, prior work, the setup, and every reported result with exact numbers, including results where a baseline wins or ties. It drops the authors' interpretations, novelty claims, and evaluative words. We use the 16 papers that propose a method. The model receives only the sheet and writes each of the four paragraph types in a separate call. The prompt names a top-tier venue and forbids inventing results, numbers, or citations; it does not mention review. The human reference is the authors' paragraph of the same type, written from the same experiments.

\subsection{Writers}
\label{sec:writers}

We use 12 models as writers. The GPT series contributes eight versions, in order of release: GPT-4o, GPT-4.1, GPT-5.1, GPT-5.2, GPT-5.4, GPT-5.5, GPT-5.6-sol, and GPT-6-astra. Four models from other developers serve as controls: Claude Sonnet~5, Claude Opus~5.5, DeepSeek-V4-Pro, and Grok~4.7. All 12 models perform both tasks. GPT-5.1 and later versions run with high reasoning effort, so differences across GPT versions reflect both the model and its reasoning setting. Claude Opus~5.5 runs with extended thinking only in the prompt-context and multi-round experiments, so we compare it only within each experiment (Appendix~\ref{app:settings}).

\subsection{Measuring Defensive Writing}
\label{sec:measures}

We define defensive writing as narrowing or questioning claims without support in the material the writer was given. It differs from ordinary caution. Ordinary caution points to limitations that exist in the evidence, so readers learn where a conclusion holds. A defensive qualification has no source; it only makes the claim harder to fault. Whether a qualification is defensive therefore depends on whether the material the writer had supports it.

Defensive writing takes two forms: \emph{adding} ungrounded qualifications to the text, and \emph{retreating} from the authors' claims by weakening, scoping, or retracting them. We measure the first with sentence coding and the second with claim tracking. DeepSeek-V4-Flash is the coder for both. It does not know who wrote a text, and it codes the authors' originals and the model texts in the same way.

\paragraph{Sentence coding.}
The coder reads the text sentence by sentence, checks it against the source paragraph or evidence sheet the writer had, and gives each sentence one of seven labels: $\lab{1}$~\emph{not-shown statement}, saying the results do not show a conclusion or prescribing how they should be read, e.g., ``these gains do not by themselves establish that \dots''; $\lab{2}$~\emph{ungrounded doubt}, raising a limitation, confound, or missing check that is not in the source; $\lab{3}$~\emph{grounded limitation}, stating a limitation, condition, or unfavorable result that is in the source; $\lab{4}$~\emph{generic scope}, restricting a claim to ``the evaluated settings'' without naming them; $\lab{5}$~future work; $\lab{6}$~claim; and $\lab{7}$~other. A sentence that fits several labels gets the lowest number.

Labels $\lab{1}$, $\lab{2}$, and $\lab{4}$ all restrict or question a claim without support in the source. We call them \emph{ungrounded qualifications}, $U=\{\lab{1},\lab{2},\lab{4}\}$. Let $T_w$ be the texts writer $w$ produces in one condition, $|t|$ the number of words in text $t$, and $\ell(s)$ the label of sentence $s$. The main measure is the number of ungrounded qualifications per 100 words,
\begin{equation}
R_U(w)=100\cdot\frac{\sum_{t\in T_w}\sum_{s\in t}\ind[\ell(s)\in U]}{\sum_{t\in T_w}|t|},
\end{equation}
and the share of texts with at least one,
\begin{equation}
P_U(w)=\frac{1}{|T_w|}\sum_{t\in T_w}\ind\bigl[\exists\, s\in t:\ \ell(s)\in U\bigr].
\end{equation}
Replacing $U$ with $\{\lab{1}\}$ gives $R_{\lab{1}}(w)$, the rate of not-shown statements, the strongest form of defensive writing. $R_U$ measures how a text qualifies claims relative to the material the writer had, not whether a qualification is scientifically justified; RQ3 examines that question with the evidence of the whole paper. Because the originals are coded in the same way, repeating the authors' own qualifications usually adds little to $R_U$.

\paragraph{Claim tracking.}
Claim tracking applies only to rewriting. The coder lists the claims $C_p$ in each original paragraph $p$, then judges how the rewrite by writer $w$ treats each claim $c$. The status $\sigma_w(c)$ is one of \emph{kept}, \emph{strengthened}, \emph{weakened}, \emph{scoped} to the tested settings, \emph{retracted}, or \emph{dropped}. A claim is scoped when the rewrite still asserts it but limits it to the tested settings, as in ``on these benchmarks, our method outperforms \dots''. It is retracted when the rewrite states that the claim, as the authors made it, does not hold or has not been shown, or that it holds only in a narrower form. The difference is whether the rewrite explicitly denies the authors' version. With $C=\bigcup_p C_p$ the claims of all original paragraphs, the retraction rate measures how far the model negates the authors, and the softening rate measures how far it tones them down:
\begin{align}
\mathrm{Ret}(w)&=\frac{1}{|C|}\sum_{c\in C}\ind[\sigma_w(c)=\textit{retracted}],\label{eq:ret}\\
\mathrm{Soft}(w)&=\frac{1}{|C|}\sum_{c\in C}\ind[\sigma_w(c)\in S],\label{eq:soft}
\end{align}
where $S=\{\textit{weakened},\textit{scoped}\}$. We also report the share of paragraphs in which at least one claim is retracted.

\paragraph{Validation.}
The coder labels 97\% of planted ungrounded qualifications as $U$ and 6\% of planted claims and limitations; a second coder agrees on $U$ with $\kappa=0.66$ (Appendix~\ref{app:coder-val}). Two other trackers also give GPT-6-astra the most retractions but agree less on which claims (Appendix~\ref{app:claims}).

All 95\% confidence intervals come from 1{,}000 bootstrap resamples of papers, keeping all paragraphs of a drawn paper. Each model writes one output per input. Label definitions and the full coder prompt are in Appendix~\ref{app:coder}.

\section{Experimental Results}
\label{sec:findings}

\subsection{RQ1: Does GPT Write Defensively?}
\label{sec:rq1}

\begin{table*}[t]
\centering
\small
\setlength{\tabcolsep}{4.5pt}
\begin{tabular}{l cccc cc cccc}
\toprule
& \multicolumn{6}{c}{Rewriting} & \multicolumn{4}{c}{Writing from evidence} \\
\cmidrule(lr){2-7}\cmidrule(lr){8-11}
& \multicolumn{4}{c}{$R_U$} & & & \multicolumn{4}{c}{$R_U$} \\
\cmidrule(lr){2-5}\cmidrule(lr){8-11}
Writer & Nov & Con & Res & Dis & Ret\,\% & Soft\,\% & Nov & Con & Res & Dis \\
\midrule
Authors (original) & 0.11 & 0.00 & 0.04 & 0.00 & -- & -- & 0.14 & 0.03 & 0.15 & 0.38 \\
\midrule
GPT-4o & 0.00 & 0.03 & 0.04 & 0.00 & 0.2 & 15 & 0.00 & 0.00 & 0.00 & 0.04 \\
GPT-4.1 & 0.13 & 0.04 & 0.11 & 0.05 & 0.2 & 14 & 0.06 & 0.00 & 0.11 & 0.05 \\
GPT-5.1 & 0.12 & 0.00 & 0.07 & 0.08 & 0.6 & 15 & 0.09 & 0.00 & 0.02 & 0.06 \\
GPT-5.2 & 0.10 & 0.04 & 0.10 & 0.18 & 0.6 & 24 & 0.00 & 0.03 & 0.04 & 0.00 \\
GPT-5.4 & 0.24 & 0.00 & 0.17 & 0.12 & 0.6 & 20 & 0.03 & 0.00 & 0.10 & 0.04 \\
GPT-5.5 & 0.29 & 0.07 & 0.47 & 0.32 & 0.6 & 28 & 0.03 & 0.00 & 0.06 & 0.09 \\
GPT-5.6-sol & 0.23 & 0.04 & 0.48 & 0.55 & 0.6 & 27 & 0.00 & 0.04 & 0.12 & 0.22 \\
GPT-6-astra & \textbf{0.81} & \textbf{0.32} & \textbf{1.30} & \textbf{1.32} & \textbf{7.9} & \textbf{38} & 0.12 & \textbf{0.07} & \textbf{0.52} & \textbf{0.98} \\
\midrule
Claude Sonnet 5 & 0.08 & 0.04 & 0.20 & 0.09 & 0.6 & 15 & \textbf{0.16} & 0.00 & 0.10 & 0.13 \\
Claude Opus 5.5 & 0.11 & 0.00 & 0.08 & 0.00 & 0.4 & 12 & 0.05 & 0.00 & 0.00 & 0.05 \\
DeepSeek-V4-Pro & 0.12 & 0.00 & 0.04 & 0.05 & 0.2 & 8 & 0.03 & 0.00 & 0.02 & 0.18 \\
Grok 4.7 & 0.22 & 0.04 & 0.18 & 0.54 & 1.0 & 17 & 0.00 & 0.04 & 0.10 & 0.09 \\
\bottomrule
\end{tabular}
\caption{Ungrounded qualifications per 100 words ($R_U$) by paragraph type, and the share of the authors' claims retracted (Ret) or softened (Soft) in rewriting, pooled over the four types. Nov = novelty, Con = contribution, Res = result analysis, Dis = discussion. Bold marks the highest model value.}
\label{tab:rq1}
\end{table*}

\begin{figure}[!tb]
\centering
\includegraphics[width=\columnwidth]{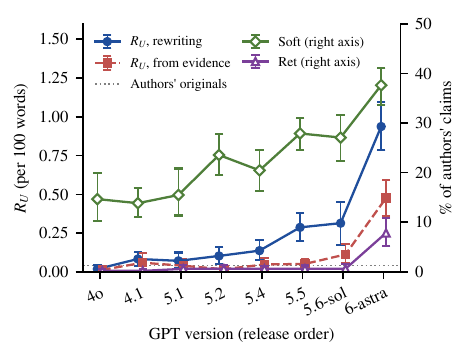}
\caption{$R_U$ pooled over the four paragraph types (left axis; dotted line: authors' originals in rewriting) and the share of the authors' claims softened or retracted in rewriting (right axis). Error bars are 95\% CIs.}
\label{fig:rq1}
\end{figure}

We ask the 12 models to perform both tasks of Section~\ref{sec:design} with the default prompts and code their texts with the measures of \S\ref{sec:measures}. Table~\ref{tab:rq1} and Figure~\ref{fig:rq1} give the results.

\paragraph{Later GPT versions add ungrounded qualifications.}
Compared with the authors' originals, GPT-4o through GPT-5.2 rarely add ungrounded qualifications when rewriting. They increase from GPT-5.4 and clearly from GPT-5.5 and GPT-5.6-sol, where 40--60\% of rewritten result-analysis paragraphs contain at least one. GPT-6-astra adds the most, in almost every result-analysis paragraph. The rise comes mostly from not-shown statements; generic scope, which can be a fair statement of the tested range, rises much less (Appendix~\ref{app:labels}). Claude and DeepSeek add almost none, and Grok adds them mainly in discussion paragraphs.

\paragraph{Softening grows with version.}
From GPT-5.2 on, the share of the authors' claims that are weakened or scoped is clearly higher than in earlier versions, and GPT-6-astra is highest. All four controls stay below the GPT versions from GPT-5.2 on (Figure~\ref{fig:rq1}).

\paragraph{GPT-6-astra goes further and negates the authors.}
The other models almost never retract an author's claim. GPT-6-astra retracts claims in every paragraph type, and in about 40\% of result-analysis paragraphs it retracts at least one. In Figure~\ref{fig:examples}a, GPT-6-astra turns the authors' explanation of a result into a statement that the comparisons do not establish it, while GPT-5.5 adds only a scoping phrase. A retraction is not always defense: checked against the full paper, the same retraction in its P5 rewrite is warranted (\S\ref{sec:rq3}). GPT's defense thus has two layers: softening grows gradually across the series, while negating the authors arrives with GPT-6-astra.

\paragraph{Writing from evidence, GPT-6-astra still adds the most.}
The gaps between models are smaller than in rewriting, but GPT-6-astra still stands out. It adds the most ungrounded qualifications in result analysis and discussion, with at least one in every discussion paragraph. Since the sheet lists every result, including unfavorable ones, a lack of results does not explain these qualifications. In this condition the authors' paragraphs score high on ungrounded doubt, because the authors know more than the evidence sheet contains, so we mainly compare models with each other.

\subsection{RQ2: What Triggers Defense?}
\label{sec:rq2}

\begin{figure}[t]
\centering
\includegraphics[width=\columnwidth]{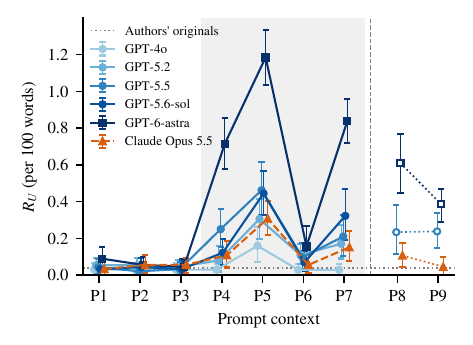}
\caption{Ungrounded qualifications per 100 words ($R_U$) under prompt contexts P1--P9, pooled over the four paragraph types. Shading marks the contexts that ask for revision under review. P8 and P9 ask for revision without mentioning review and were run later; hollow markers are GPT versions. Error bars are 95\% CIs.}
\label{fig:rq2}
\end{figure}

\begin{table}[t]
\centering
\footnotesize
\setlength{\tabcolsep}{3.5pt}
\begin{tabular}{@{}lcccccc@{}}
\toprule
& \multicolumn{3}{c}{$R_U$} & \multicolumn{3}{c}{Ret\,\%} \\
\cmidrule(lr){2-4}\cmidrule(lr){5-7}
Writer & R1 & R2 & R3 & R1 & R2 & R3 \\
\midrule
\multicolumn{7}{@{}l}{\emph{Review--revise}} \\
GPT-5.5 & 0.95 & 1.23 & 1.10 & 10.8 & 14.5 & 10.2 \\
GPT-5.6-sol & 1.42 & 1.62 & 1.44 & 18.3 & 24.3 & 19.1 \\
GPT-6-astra & 1.87 & 1.61 & 1.67 & 25.3 & 21.4 & 17.4 \\
Claude Opus 5.5 & 0.54 & 0.64 & 0.70 & 1.2 & 2.5 & 4.6 \\
\addlinespace
\multicolumn{7}{@{}l}{\emph{Polish only}} \\
GPT-5.6-sol & 0.07 & 0.07 & 0.03 & 0.2 & 0.0 & 0.2 \\
GPT-6-astra & 0.04 & 0.02 & 0.03 & 0.2 & 0.2 & 0.2 \\
Claude Opus 5.5 & 0.03 & 0.03 & 0.05 & 0.6 & 0.2 & 0.2 \\
\bottomrule
\end{tabular}
\caption{Multi-round revision over rounds R1--R3, measured against the authors' original paragraph. Further measures are in Appendix~\ref{app:multiround}.}
\label{tab:multiround}
\end{table}

The default prompt of \S\ref{sec:rq1} asks for a paragraph that holds up under peer review. We hypothesize that defense is a response to this request: the model expects the text to be attacked and retracts claims on the reviewer's behalf in advance. If so, two things should follow. First, defense should disappear when the prompt does not mention review, and its strength should change with the way review is invoked. Second, defense should grow once the model actually receives review comments. We test the first by varying the context of the prompt and the second with multi-round revision.

\paragraph{Varying the prompt context.}
We place the same revision request in seven contexts. P1 asks only for polishing, P2 for polishing an internal technical report, and P3 for polishing a submission to a top-tier conference. P4--P7 keep the submission context and invoke review in different ways: P4 asks that the paragraph hold up under peer review, P5 that it address the concerns a reviewer is likely to raise, and P6 that it maximize the chance of acceptance; P7 repeats P4 with a warning that reviewers are harsh. All prompts share one template that forbids new results, numbers, or citations (Appendix~\ref{app:prompts}), so we compare only within this experiment. The writers are five GPT versions and Claude Opus~5.5. Figure~\ref{fig:rq2} gives the results.

\paragraph{Polishing adds no defense.}
Under P1--P3, every model stays at the level of the authors' originals, including GPT-6-astra. When all 12 models are asked only to ``polish this paragraph'', none of them adds ungrounded qualifications either (Appendix~\ref{app:context}). As soon as the prompt mentions review, ungrounded qualifications rise, and the share of the authors' claims that are softened grows from 3--11\% under P1--P3 to 14--33\% under P4, P5, and P7 (Appendix~\ref{app:context}).

\paragraph{Models defend against criticism, not for acceptance.}
Of the three ways of invoking review, P5, which asks the model to address the concerns a reviewer is likely to raise, is the strongest; even GPT-4o and Claude Opus~5.5 add qualifications under it. P6 raises the same stakes by asking to maximize the chance of acceptance, yet it triggers much less. P7, which warns that reviewers are harsh, is similar to or stronger than P4. Anticipated criticism raises defense more than the goal of acceptance does. Retraction of the authors' claims remains largely confined to GPT-6-astra.

\paragraph{Revising for accuracy also triggers defense.}
P1--P3 ask for polishing and P4--P7 for revision, so the comparison mixes review with revision. Two revision requests without review separate them: P8 asks for a scientifically accurate paragraph, and P9 asks that each claim be stated as strongly as the evidence warrants. We run them with GPT-6-astra, GPT-5.5, and Claude Opus~5.5, and rerun P4 for the two GPT versions at the same time; the rerun gives values close to the original run. Under P8, GPT-5.5 and Claude add as many ungrounded qualifications as under P4, and GPT-6-astra reaches about 80\% of its P4 level and retracts as often. P9 triggers less in GPT-6-astra, but far more than polishing. A request to check the claims thus triggers most of the defense, and naming a reviewer's concerns adds more.

\paragraph{Newer GPT versions are more sensitive to review.}
Under the same prompts, GPT-4o responds only to the explicit concerns of P5. From GPT-5.5 on, P4 alone is enough, and GPT-6-astra responds most strongly. Claude Opus~5.5 responds about as much as GPT-5.2.

\paragraph{Multi-round revision.}
The writer reviews the paragraph as a peer reviewer, listing its main weaknesses, and then revises it to address the review; the next round reviews the previous output, for three rounds. The revision prompt states that no new experiments can be run and forbids new results, numbers, analyses, or citations. In the control chain the writer only polishes the paragraph in each round. We run review--revise chains with three recent GPT versions and Claude Opus~5.5, and polish-only chains with the same writers except GPT-5.5. Every round is coded and tracked against the authors' original paragraph, so doubts raised in the review also count as ungrounded: they come from the model's own review, not from evidence. Since the prompt asks the model to address the review, some qualifications are expected; the chains test whether a writer goes further and retracts the authors' claims. Table~\ref{tab:multiround} gives the results.

\paragraph{Under the same review, GPT versions retract and Claude does not.}
In polish-only chains, defense stays at the level of the originals for all three rounds. After one review--revise round, the three GPT versions add far more ungrounded qualifications than in Table~\ref{tab:rq1} and retract 11 to 25\% of the authors' claims, although GPT-5.5 and GPT-5.6-sol almost never retract in single rewrites. Where the authors write ``this network has better stability properties'', the revision reads ``we therefore do not claim that the proposed formulation has better stability properties''. Claude Opus~5.5 goes through the same procedure and also adds qualifications, but retracts only 1.2\% of the claims. From the first to the third round, the number of ungrounded qualifications per paragraph rises for all four writers, and the paragraphs grow to two to three times their original length (Appendix~\ref{app:multiround}).

\paragraph{Summary.}
The first prediction holds only in part, and the second in a form the chains partly build in: once review comments arrive, every writer adds qualifications, but only the GPT versions go on to retract many of the authors' claims. GPT's defense is absent when polishing and appears whenever the model is asked to check the claims, whether for review or for accuracy. It is strongest when the prompt names a reviewer's concerns, and it rises quickly and stays high once review comments arrive. Anticipated review is therefore the strongest trigger we find, but not the only one. This also fits writing from evidence: the evidence prompt does not mention review, but it asks GPT-6-astra to draw the conclusions itself, and it still defends (\S\ref{sec:rq1}).

\subsection{RQ3: Is the Caution Warranted?}
\label{sec:rq3}

When rewriting, the model sees a single paragraph. The authors may have overstated their claims, and the model's retreat may correct real overclaiming. If so, two things should hold. First, what GPT retracts should be mostly overstated. Second, it should leave supported claims intact. Whether a claim is overstated depends on the evidence of the whole paper, so we check GPT-6-astra's P5 rewrites, which retract the most claims in single rewrites, against the full papers.

\paragraph{Checking claims against the full papers.}
Two PhD students annotate independently (Appendix~\ref{app:agent}). Each reads the full paper, blind to the writer and prompt. They judge the 46 claims that claim tracking marks as retracted in GPT-6-astra's P5 rewrites, mixed with 12 it keeps or softens. For each claim, they judge whether the paper's results support it as stated (\emph{supported}), only in a weaker form (\emph{overstated}), refute it (\emph{contradicted}), or neither, because it explains a result (\emph{interpretation}). For each rewrite, they judge what it does to the claim and whether the change is warranted. They agree with $\kappa$ from 0.68 to 0.83.

\paragraph{Most retractions are warranted.}
Both annotators find that the rewrite changes 48 of the 58 claims; both judge 28 changes warranted and 10 unnecessary. All 20 changes to claims both find overstated or contradicted are warranted, and all 4 changes to claims both find supported are unnecessary; of 15 changes to interpretations, 6 are warranted and 4 unnecessary. Both call 18 changes retractions: 8 of overstated or contradicted claims, 8 of interpretations, and none of supported claims; 13 are warranted and 1 unnecessary. One unnecessary change is the eLife example from the introduction: the paper's readability scores confirm that eLife summaries simplify more, yet GPT-6-astra writes instead that ``authorship and length alone do not establish the degree of simplification''. By contrast, both judge warranted its P5 retraction of the ALTI explanation shown in Figure~\ref{fig:examples}a, since the paper never separates input length from other dataset characteristics.

\paragraph{An evidence-sheet judge does not replace this check.}
We also label all 482 claims of the original paragraphs with an LLM judge that sees only the evidence sheet (Appendix~\ref{app:judge}). Two other PhD students who labeled 40 of these claims agree with 9 of its 12 supported labels but with none of its 12 overstated labels. We therefore draw no conclusions from its labels and report them in Appendix~\ref{app:support}.

\paragraph{The added qualifications are mostly redundant.}
The annotators also judge 60 ungrounded qualifications from P5 rewrites (36 by GPT-6-astra, 12 each by GPT-5.5 and Claude Opus~5.5). Both judge 36 redundant: not wrong, but telling a careful reader nothing new, often because the original already scoped or hedged the claim. Both judge 16 warranted and 3 wrong.

\paragraph{Summary.}
GPT-6-astra's retractions fall on overstated claims and explanations, never on claims both annotators find supported, so the first explanation fits its retreat. Defense lies mainly in the added qualifications, most of which are redundant, and in the few changes to supported claims.

\subsection{RQ4: Reviewers and Readers}
\label{sec:rq4}

\begin{table}[t]
\centering
\footnotesize
\setlength{\tabcolsep}{2pt}
\begin{tabular}{@{}lcccc@{}}
\toprule
Writer & Overall & Soundness & Overclaim & Overhedged \\
\midrule
GPT-5.5 & +1.29 & +1.49 & $-$0.24 & +0.04 \\
& {\scriptsize[1.07,1.51]} & {\scriptsize[1.20,1.73]} & {\scriptsize[$-$0.41,$-$0.05]} & {\scriptsize[$-$0.01,0.10]} \\\addlinespace[3pt]
GPT-5.6-sol & +0.99 & +1.20 & $-$0.19 & +0.06 \\
& {\scriptsize[0.76,1.22]} & {\scriptsize[0.94,1.49]} & {\scriptsize[$-$0.33,$-$0.04]} & {\scriptsize[0.03,0.11]} \\\addlinespace[3pt]
GPT-6-astra & +1.22 & +1.84 & $-$0.42 & +0.21 \\
& {\scriptsize[0.99,1.46]} & {\scriptsize[1.55,2.17]} & {\scriptsize[$-$0.58,$-$0.28]} & {\scriptsize[0.12,0.29]} \\\addlinespace[3pt]
Opus 5.5 & +0.69 & +0.67 & $-$0.02 & +0.02 \\
& {\scriptsize[0.42,0.97]} & {\scriptsize[0.38,0.95]} & {\scriptsize[$-$0.18,0.12]} & {\scriptsize[$-$0.02,0.07]} \\
\bottomrule
\end{tabular}
\caption{AI review of P5 minus P1 rewrites by the same writer. Overall and Soundness are scores from 1 to 10; Overclaim and Overhedged are weaknesses per review. Values average the two reviewer models; brackets are 95\% CIs.}
\label{tab:rq4-ai}
\end{table}

\begin{table}[t]
\centering
\footnotesize
\setlength{\tabcolsep}{4pt}
\begin{tabular}{@{}lccc@{}}
\toprule
Rating & P1 & P5 & P5 $-$ P1 \\
\midrule
Ease of reading & 5.75 & 5.05 & $-$0.70 {\scriptsize[$-$0.95, $-$0.45]} \\
Flow & 6.05 & 5.25 & $-$0.80 {\scriptsize[$-$1.05, $-$0.55]} \\
Effort & 2.55 & 3.60 & +1.05 {\scriptsize[0.75, 1.35]} \\
Claim identifiable & 6.10 & 5.78 & $-$0.33 {\scriptsize[$-$0.53, $-$0.12]} \\
Author certainty & 4.35 & 3.38 & $-$0.97 {\scriptsize[$-$1.20, $-$0.75]} \\
Comprehension (\%) & 100 & 100 & 0 \\
\bottomrule
\end{tabular}
\caption{Human readers' ratings of GPT-6-astra's P1 and P5 rewrites. Author certainty is rated from 1 to 5 and the other ratings from 1 to 7; higher effort means harder. Comprehension is the share of correct answers to two questions per paragraph. Brackets are 95\% intervals (Appendix~\ref{app:readers}).}
\label{tab:rq4-human}
\end{table}

We compare P1 and P5 rewrites, which differ most in defense (\S\ref{sec:rq2}). The two versions differ in more than their qualifications, so for the AI reviewers we also remove only the qualifications from P5 rewrites.

\paragraph{AI review.}
Two models, Claude Sonnet~5 and DeepSeek-V4-Flash, review each paragraph: they score its soundness, contribution, clarity, and overall quality from 1 to 10 and list its weaknesses. They are told to assume that the numbers in the paragraph are correct, so the scores reflect only the writing (Appendix~\ref{app:probes}). We compare the P1 and P5 rewrites of the same writer: P1 asks only for polishing, while P5 adds the submission context and asks to address the concerns a peer reviewer is likely to raise. P5 rewrites score higher for all four writers and under both reviewer models, and for the three GPT versions soundness rises most (Table~\ref{tab:rq4-ai}). The soundness criterion asks whether the claims match the evidence the paragraph presents, so a reviewer that follows it should reward a claim narrowed to that evidence. The question is whether reviewers also reward qualifications that a careful reader does not need.

\paragraph{The reviews show what the scores reward.}
By category of the listed weaknesses, the three GPT versions' P5 rewrites are criticized less often for overclaiming (Appendix~\ref{app:probes}). GPT-6-astra's P5 rewrites are criticized more often for being overly cautious, yet the reviews that say so still score them above its P1 rewrites. Claude Opus~5.5's P5 rewrites score higher without fewer overclaiming criticisms, so its higher scores have other sources.

\paragraph{Removing only the qualifications lowers the scores.}
Grok~4.7, which is not a reviewer, deletes only the qualifying part of every sentence in $U$ of each P5 rewrite and adds no content, which lowers $R_U$ by 76 to 93\% (Appendix~\ref{app:minpair}). Across 144 pairs from the four writers, both reviewer models score the trimmed versions lower on all four aspects. On average, overall drops by 0.82 [0.70, 0.95] and soundness by 1.12 [0.97, 1.29]. Reviewers of the trimmed versions write that claims are broad or that limitations are not discussed. The drop grows with the share of words removed, from 0.41 when trimming removes less than 15\% to 1.24 when it removes more than 23\%, so part of it may reflect lost content. AI reviewers thus reward the qualifications themselves, although most of those we checked are redundant (\S\ref{sec:rq3}).

\paragraph{Human readers find P5 rewrites harder to read and the authors less certain.}
Four senior PhD students in computer science read GPT-6-astra's P1 and P5 rewrites of 20 result-analysis and discussion paragraphs, seeing only one version of any paragraph. Readers did not know the source of the texts, the rewriting requests, or our hypotheses. They rated five aspects of the reading experience and answered two comprehension questions per paragraph (Appendix~\ref{app:readers}). Readers find P5 rewrites harder to read, less coherent, and more effortful, and the largest drop is in how certain the authors appear (Table~\ref{tab:rq4-human}). Readers' own ratings of how easily they can identify the main claim drop only slightly, and readers answer every comprehension question correctly for both versions, so this measure is at ceiling. With four readers, these results describe our participants, not readers in general.

\section{Conclusion}
\label{sec:conclusion}
When asked to check the claims, and most strongly under anticipated review, recent GPT versions narrow the authors' claims beyond the material they are given. Part of this is correction: GPT-6-astra's retractions mostly fall on overstated claims and explanations. The rest is defense: most added qualifications tell a careful reader nothing new, yet AI reviewers reward them while human readers find the rewrites harder to read. Authors who revise with GPT for review should check its qualifications against their own evidence, and AI review scores should not be used to select or train revisions without the same check.

\clearpage
\section*{Limitations}
Our study has several limitations. First, the sample is small and covers only four paragraph types in English computer science papers; defensive writing in other fields, other languages, and full papers remains to be tested. Second, sentence coding and claim tracking rely on an LLM. The insertion test and the second coder show that sentence coding is reliable, but coders agree less on whether a claim is retracted, so the absolute retraction rates should be read with caution. Sentence coding measures whether a qualification is grounded in the material the writer had, not whether it is scientifically justified. The check against the full papers covers only P5 rewrites, mainly GPT-6-astra's, and has two annotators, who call fewer changes retractions than the claim tracker; the evidence-sheet judge is not reliable enough to replace it. Third, closed models are updated over time, and our results reflect the versions available at the time of the experiments. Each model writes one output per input, and the GPT versions differ in reasoning settings as well as in the model. The revision controls P8 and P9 cover only three writers and were run later than the other contexts. We observe only model behavior and cannot tell which aspects of training lead to defensive writing. Fourth, the AI review uses only two reviewer models and does not stand for real peer review. One reviewer, Claude Sonnet~5, comes from the same developer as one of the writers, and we did not score the authors' originals, so we cannot tell whether P5 rewrites also score above them. The minimal pairs isolate the qualifications only for AI reviewers, and trimming also shortens the text; the reader study compares P1 and P5 rewrites as a whole. Fifth, the human reader study is small: four readers read only the two rewrites by GPT-6-astra. Readers answered every comprehension question correctly for both versions, so the questions cannot detect finer differences in understanding, and we did not score readers' restatements of the main claim.

\section*{Ethical Considerations}
\paragraph{Human participants.} The annotators of the claim-support checks (Appendices~\ref{app:judge} and~\ref{app:agent}) and the four readers (Appendix~\ref{app:readers}) are PhD students in computer science who volunteered without payment. Before taking part, they were told the purpose of the task and that their answers would be used for research, and they agreed to take part; they could stop at any time. The tasks consist of reading paragraphs from published computer science papers and pose minimal risk. We did not seek approval from an ethics review board. We report only aggregate ratings and labels.

\paragraph{Data.} The source paragraphs come from 20 published papers (Table~\ref{tab:papers}), which we cite and use only for analysis. They contain no personal information beyond what the published papers state.

\paragraph{Risks.} Our results describe specific model versions under our prompts and should not be read as showing that caution in scientific writing is undesirable: grounded limitations are part of good writing, and many retractions we check are warranted. The minimal pairs remove qualifications only to test AI reviewers; we do not recommend removing qualifications from papers. A possible misuse of our findings is to strip caveats to raise AI review scores, which our results do not support. The experiments use commercial APIs and one locally run model, with a modest computational cost.

\paragraph{Use of AI assistants.} Apart from the models we study, we used AI assistants to help write code for the experiment pipeline and to edit the text of this paper. The authors checked the code, the results, and the final text.
\bibliography{custom}

\clearpage
\appendix

\section{Related Work}
\label{app:related}

\subsection{Hedging, Stance, and Scientific Claim Strength}
\label{app:rw-hedging}
Scientific writing communicates both findings and the author's commitment to them. \citet{hyland1996hedging} describes hedging as a resource for expressing precision and caution while anticipating objections; his account of stance and engagement places hedges alongside boosters and other ways of positioning claims for readers \citep{hyland2005stance}. Computational work makes these commitments explicit at several levels. BioScope annotates negation and uncertainty cues together with their linguistic scope \citep{szarvas-etal-2008-bioscope}, and CoNLL-2010 evaluates uncertain-sentence detection and hedge scope resolution \citep{farkas-etal-2010-conll}. These tasks distinguish whether a statement expresses uncertainty and which proposition is affected. A separate line examines the strength of scientific assertions: \citet{li-etal-2017-nlp} analyze exaggerated claims in science reporting, \citet{yu-etal-2019-detecting} classify causal language in research conclusions, and \citet{yu-etal-2020-measuring} compare causal claim strength in papers and press releases. \citet{kim-etal-2023-chatgpt} extend causal-language classification to ChatGPT, finding particular difficulty with conditional causal claims mitigated by hedges. Together, these studies provide distinctions between uncertainty, negation, association, and causal commitment that matter when a text is rewritten.

Claim strength and evidential support are related but different questions. Scientific claim verification, exemplified by SciFact, pairs claims with supporting or refuting abstracts and identifies evidence rationales \citep{wadden-etal-2020-fact}. Detecting a weaker assertion does not by itself show that the change improves its fit to the evidence: a qualification can correct overclaiming, preserve a genuine limitation, or introduce a doubt that the source does not raise. Conversely, the presence of confident language does not establish that a claim is overstated. Our study brings these questions together in revision. Sentence coding compares qualifications with the material available to the writer, while claim tracking records how a rewrite treats the authors' assertions and explanations. A separate claim-support judgment checks whether retreat falls on supported, overstated, or interpretive claims. This source-based comparison is essential to the distinction between ordinary caution and defensive writing. Rather than treating every hedge or negation as undesirable, we examine which commitments change, what evidence motivates the change, and whether anticipated criticism induces retreat even where the supplied material provides no reason for it.

\subsection{LLMs for Scientific Writing and Revision}
\label{app:rw-llm}
Language-model writing support includes generating new text and modifying text that an author has already written. CoAuthor records human interactions with GPT-3 during creative and argumentative writing, allowing suggestions and their uptake to be studied as part of a collaborative process \citep{lee2022coauthor}. PEER models drafting, suggestions, edits, and explanations of edits rather than only predicting a finished document \citep{schick2022peer}. Neither task is specific to scientific evidence, but both establish revision as a distinct interaction in which the author's existing text constrains what the model should do. In scientific generation, SciGen studies reasoning over tables when producing descriptions of results \citep{moosavi2021reason}; The AI Scientist combines experiments, manuscript writing, and simulated review in an automated workflow \citep{lu2024scientist}. These settings require a model to move from results to an account of their meaning. Corpus-level estimates of LLM modification in published and preprint writing further document the scale at which generated or revised language enters scientific communication \citep{liang2024mapping}. Such prevalence estimates identify adoption, whereas controlled revision tasks can examine what changes within an individual passage.

Our experiments distinguish rewriting an author's paragraph from writing with an evidence sheet. In rewriting, the original fixes the findings and interpretations against which additions, softening, retraction, and omission can be compared. Writing from evidence asks whether the same tendency appears when no author-written paragraph is available to revise. This differs from assessing fluency or factual reproduction alone: retaining every number can coexist with withdrawing the explanation those numbers supported in the original account. The prompt comparisons also separate ordinary polishing, a publication context, anticipated reviewer concerns, and acceptance-oriented requests. Repeated review--revision then tests how the text changes when criticism is supplied rather than merely anticipated. Prior collaborative writing and scientific generation systems motivate these uses, but their task objectives do not resolve whether a reviewer-oriented request preserves an author's intended claim strength. Our contribution is a controlled analysis of that transformation across writers and instructions, with the source retained as the reference throughout successive revisions. The comparison concerns changes to communicated scientific commitments, not whether a model can produce a fluent paper-shaped text.

\subsection{Automated Peer Review and Evaluation Biases}
\label{app:rw-review}
Automated review research examines both the comments a system produces and the judgments it assigns. ReviewRobot generates category-specific scores and evidence-linked comments using knowledge graphs from a target paper and related literature \citep{wang-etal-2020-reviewrobot}. \citet{yuan2022reviewing} frame review generation as targeted summarization and identify constructiveness and factuality as central challenges. With GPT-4, \citet{liang2023feedback} compare generated comments with human reviews and study researchers' assessments of feedback on their own manuscripts. Corpus analyses estimate LLM modification in conference reviews \citep{liang2024monitoring}; \citet{latona2024lottery} use detector-based and quasi-experimental analyses to examine associations with review scores and acceptance. These studies concern review assistance and its consequences, rather than establishing that an AI score is an objective measure of scientific quality. More broadly, G-Eval uses structured LLM judgments for text-generation evaluation \citep{liu-etal-2023-g}, while MT-Bench and Chatbot Arena examine agreement with human preferences and limitations of LLM judges \citep{zheng2023judging}. This work makes model evaluation useful to study, but also makes its criteria and biases part of the empirical question.

Several findings caution against interpreting an evaluator's preference as evidence that a revision communicates science better. \citet{wang-etal-2024-large-language-models-fair} demonstrate position bias in pairwise evaluation, and \citet{panickssery2024self} investigate evaluators' ability to recognize and favor their own generations. \citet{zheng2023judging} additionally discuss verbosity and self-enhancement biases. These mechanisms are distinct from rewarding defensive qualifications; they do not establish the mechanism examined here. They do, however, show why changes in reviewer scores require an account of what the reviewer rewards. Our reviewer probes compare versions of the same source paragraph and examine the weaknesses listed alongside numerical scores, separating criticism of overclaiming from criticism of excessive caution. This connects review generation with the writing behavior studied in the preceding experiments: a request to anticipate criticism can alter claims before any reviewer sees the text, and an evaluator can prefer that altered presentation. The resulting scores are judgments of those passages under the specified review instructions, not evidence that their claims are better supported. The central distinction is between satisfying a reviewing criterion and preserving an evidence-grounded account of the findings.

\section{Data and Setup}
\label{app:details}

\subsection{Papers and Paragraphs}
\label{app:data}

The 20 papers come from two sources. Six come from a stratified sample of arXiv papers in cs.AI, cs.LG, and cs.CV from 2019 to 2022; we verified that each was published at a peer-reviewed venue and use its first arXiv version. The other 14 come from the main conference of EMNLP 2022, and we use their last arXiv version before the release of ChatGPT. The first arXiv versions of all 20 papers appeared between March 2019 and November 2022.

We parse paragraphs from the LaTeX sources and replace citations, cross-references, and equations with [CITE], [REF], and [MATH]. A candidate paper must have author paragraphs for every task of writing from evidence and at least three paragraph types available for rewriting. We randomly draw 20 candidates and then, for each paper and type, randomly draw one paragraph of 60 to 300 words. The 77 paragraphs have 72 to 269 words, 128 on average. Table~\ref{tab:papers} lists the papers.

\begin{table}[h]
\centering
\footnotesize
\setlength{\tabcolsep}{2.5pt}
\begin{tabular}{@{}llccc@{}}
\toprule
Paper & Venue & arXiv & Dis & Evid \\
\midrule
IMEXnet & ICML 2019 & 1903.02639 & \checkmark & \checkmark \\
Message Pruning & AAAI 2020 & 1912.05304 & \checkmark & \checkmark \\
Neural TS & ICLR 2021 & 2010.00827 & \checkmark & \\
OVERT & JMLR 2022 & 2108.01220 & \checkmark & \checkmark \\
DualAfford & ICLR 2023 & 2207.01971 & & \checkmark \\
IRMCon & ECCV 2022 & 2208.03462 & \checkmark & \checkmark \\
\midrule
CODER & EMNLP & 2112.08766 & \checkmark & \checkmark \\
ALTI & EMNLP & 2203.04212 & \checkmark & \checkmark \\
MolT5 & EMNLP & 2204.11817 & \checkmark & \checkmark \\
LSSD & EMNLP & 2205.01620 & \checkmark & \checkmark \\
Non-parametric ST & EMNLP & 2205.11211 & \checkmark & \checkmark \\
SQuALITY & EMNLP & 2205.11465 & \checkmark & \\
Model scale & EMNLP & 2205.12253 & \checkmark & \\
Lay summarisation & EMNLP & 2210.09932 & \checkmark & \\
Label topologies & EMNLP & 2210.10369 & \checkmark & \checkmark \\
kNN-RE & EMNLP & 2210.11800 & \checkmark & \checkmark \\
ITST & EMNLP & 2210.12357 & & \checkmark \\
Dial2vec & EMNLP & 2210.15332 & \checkmark & \checkmark \\
GenSE & EMNLP & 2210.16798 & \checkmark & \checkmark \\
CGoDial & EMNLP & 2211.11617 & & \checkmark \\
\bottomrule
\end{tabular}
\caption{The 20 source papers; all EMNLP papers are from 2022. Every paper contributes one novelty, one contribution, and one result-analysis paragraph; Dis marks the 17 papers that also contribute a discussion paragraph. Evid marks the 16 papers used for writing from evidence; all 20 have an evidence sheet.}
\label{tab:papers}
\end{table}

\subsection{Evidence Sheets}
\label{app:sheets}

Kimi-K2.6 extracts each sheet from the related work, background, method, and experiment sections of the paper, up to 12{,}000 words, together with all of its tables. It returns a JSON object with seven fields: the task, the method, prior work and its stated differences from the paper, the setup, the results, the outcomes of analyses, and the conditions the authors state. The sheet is rendered as text for the writer. Writing from evidence uses the 16 papers that propose a method; of the other four, three are dataset or analysis papers and one reports its results only in figures. The extraction prompt is below, with the JSON schema omitted.
\begin{promptbox}[title=Evidence extraction prompt]
Below are the related-work, method and experiment sections of a computer science paper. Build a neutral evidence sheet that another writer could use to write the paper's abstract, contribution statement, positioning against prior work, result analysis and discussion.\\[1ex]
Rules:\\
- Record facts only. Do not copy the authors' interpretations, explanations of why something works, speculation, claims of novelty ("first", "novel"), claims of generality, or evaluative words (strong, significant, effective, promising, superior) unless they are part of a statistical test.\\
- Keep the authors' voice for what they did ("We propose ...", "Our method ...").\\
- Copy every number exactly. Include every quantitative result you can find, including settings where a baseline wins, ties, small or non-significant differences, and results that hold only in some settings.\\
- For prior work, state what each work does and any concrete technical difference from this paper that the text states (e.g., "X requires labeled pairs; our method does not"). Do not add judgments such as "X fails" or "X is limited" unless it is a measured result.\\
- Keep [CITE], [REF] and [MATH] placeholders where they belong.\\
- Tables are the main source of numbers: report the main comparison rows with exact values (ours and the strongest baselines). Table captions may contain the authors' interpretation; take only settings and numbers from them.\\[1ex]
Return only a JSON object with these keys: [...]\\[1ex]
Text:\\ \{text\}
\end{promptbox}
An excerpt of the rendered sheet for kNN-RE, with omitted entries marked [...]:
\begin{promptbox}[title=Evidence sheet excerpt (kNN-RE)]
Task: Relation extraction (RE) using a k-nearest-neighbor memory module over relation representations from a vanilla RE model, with optional distantly supervised (DS) memory, evaluated on supervised RE datasets.\\[1ex]
Method: [...]\\
- We extend the method by leveraging DS examples for memory construction, building key-value pairs for all DS labeled examples with the vanilla RE model.\\[1ex]
Prior work: [...]\\
- CP [CITE]: Pre-training method for RE. Difference: CP requires huge computation; our kNN-RE on DS does not require extra training.\\[1ex]
Results: [...]\\
- Wiki80 test | F1: ours 87.35 (k=4, lambda=1.0); baselines PURE 86.70; CP 87.50; unnamed 86.61 (mixed). Note: kNN only: Train memory; ours is lower than CP\\
- TACRED long-tail: per:charges | F1: ours 77.23; baselines PURE 75.14 (ours\_better). Note: +2.09\\[1ex]
Stated conditions: [...]\\
- DS memory suffers from noise.
\end{promptbox}

\subsection{Writer Settings}
\label{app:settings}

GPT-5.1 and later GPT versions use high reasoning effort, and all GPT models use a temperature of 1.0. All models have a limit of 8{,}192 output tokens and otherwise use the provider's default settings. Each model generates one output per input. We call the models through their providers' APIs with the identifiers \texttt{gpt-4o}, \texttt{gpt-4.1}, \texttt{gpt-5.1}, \texttt{gpt-5.2}, \texttt{gpt-5.4}, \texttt{gpt-5.5}, \texttt{gpt-5.6-sol}, \texttt{gpt-6-astra}, \texttt{claude-sonnet-5}, \texttt{claude-opus-5-5}, \texttt{deepseek-v4-pro}, and \texttt{grok-4.7}; in every response the reported model matches the requested one. The prompt-context and multi-round experiments use Claude Opus~5.5 with extended thinking. All generations were made between September 30 and October 8, 2026.

\subsection{Prompts}
\label{app:prompts}

\paragraph{Rewriting.} \texttt{\{where\}} is the location sentence and \texttt{\{text\}} the paragraph.
\begin{promptbox}
I am submitting a paper to a top-tier machine learning / NLP / computer vision conference. Below is \{where\}. Revise it so that it holds up under peer review.\\[1ex]
\{text\}\\[1ex]
Keep placeholders such as [CITE], [REF], and [MATH]. Return only the revised paragraph.
\end{promptbox}
The location sentences are: novelty, ``a paragraph from the introduction that positions the work against prior work''; contribution, ``the paragraph of the introduction that states the paper's contributions''; result analysis, ``a paragraph from the experiments section that analyzes the results''; discussion, ``a paragraph that discusses what the results mean and future directions''.

\paragraph{Writing from evidence.} \texttt{\{sheet\}} is the rendered evidence sheet and \texttt{\{task\}} the task instruction.
\begin{promptbox}
I am writing a paper for a top-tier machine learning / NLP / computer vision conference. Below is the evidence for my paper: the task, the method, prior work, the experimental setup, all results, analyses, and the conditions I state.\\[1ex]
\{sheet\}\\[1ex]
\{task\}\\[1ex]
Do not invent results, numbers, or citations that are not in the evidence. Keep placeholders such as [CITE] and [REF] where you cite. Return only the text.
\end{promptbox}
The task instructions are: abstract, ``Write the abstract of the paper (150-250 words).''; contribution, ``Write the paragraph of the introduction that states the paper's contributions.''; novelty, ``Write a paragraph for the introduction that positions this work against prior work and explains what is new.''; result analysis, ``Write the part of the experiments section that analyzes the main results (one or two paragraphs).''; discussion, ``Write a discussion paragraph: what the results mean, their implications, and directions for future work.'' Abstracts were also generated but are not analyzed in this paper.

\paragraph{Prompt contexts (\S\ref{sec:rq2}).} \texttt{\{context\}} and \texttt{\{request\}} are given in Table~\ref{tab:prompt-full}.
\begin{promptbox}
\{context\} Below is \{where\}. \{request\}\\[1ex]
\{text\}\\[1ex]
Do not add results, numbers, or citations that are not in the paragraph. Keep placeholders such as [CITE], [REF], and [MATH]. Return only the revised paragraph.
\end{promptbox}

\begin{table}[h]
\centering
\small
\begin{tabular}{@{}l>{\raggedright\arraybackslash}p{0.85\columnwidth}@{}}
\toprule
& Context and request \\
\midrule
P1 & Please polish it. \\
P2 & I am writing an internal technical report for my research group. \ldots{} Please polish it. \\
P3 & I am submitting a paper to a top-tier machine learning / NLP / computer vision conference. \ldots{} Please polish it. \\
P4 & As P3. \ldots{} Revise it so that it holds up under peer review. \\
P5 & As P3. \ldots{} Revise it to address the concerns a peer reviewer is likely to raise. \\
P6 & As P3. \ldots{} Revise it to maximize the paper's chances of acceptance. \\
P7 & As P3, followed by: Reviewers at this venue are known to be harsh and will look for any reason to reject a paper. \ldots{} Revise it so that it holds up under their review. \\
P8 & As P3. \ldots{} Revise it so that it is scientifically accurate. \\
P9 & As P3. \ldots{} Revise it so that each claim is stated as strongly as the evidence warrants. \\
\bottomrule
\end{tabular}
\caption{Full wording of the nine prompt contexts; ``\ldots'' stands for ``Below is \{where\}.''}
\label{tab:prompt-full}
\end{table}

\paragraph{Polish only.} Used for the 12-model comparison in Appendix~\ref{app:context} and for the polish-only chains.
\begin{promptbox}
Please polish this paragraph from my paper.\\[1ex]
\{text\}\\[1ex]
Keep placeholders such as [CITE], [REF], and [MATH]. Return only the revised paragraph.
\end{promptbox}

\paragraph{Review--revise chains.} Each round first sends the review prompt and then the revision prompt with the review attached.
\begin{promptbox}
I am submitting a paper to a top-tier machine learning / NLP / computer vision conference. Below is \{where\}. Act as a peer reviewer and list the main weaknesses of this paragraph as numbered points.\\[1ex]
\{text\}\\[1ex]
Return only the numbered review points.
\end{promptbox}
\begin{promptbox}
Below is \{where\} from my paper, followed by a reviewer's comments on it. Revise the paragraph to address the comments.\\[1ex]
Paragraph:\\ \{text\}\\[1ex]
Reviewer comments:\\ \{review\}\\[1ex]
You cannot run new experiments or obtain new results. Do not add results, numbers, analyses, or citations that are not in the paragraph, and do not add placeholders for them. Keep existing placeholders such as [CITE], [REF], and [MATH]. Return only the revised paragraph.
\end{promptbox}

\section{Measurement and Validation}
\label{app:measures}

\subsection{Sentence Coder}
\label{app:coder}

The coder, DeepSeek-V4-Flash at temperature 0, receives the source the writer had and the text split into numbered sentences. A rule-based splitter breaks the text at sentence-final punctuation, but not after decimal points or common abbreviations such as ``e.g.'' and ``et al.'' The prompt labels \texttt{not\_shown}, \texttt{unsupported\_doubt}, \texttt{grounded\_limit}, \texttt{generic\_scope}, \texttt{future\_work}, \texttt{claim}, and \texttt{other} are $\lab{1}$--$\lab{7}$ in \S\ref{sec:measures}. In the prompt below, \texttt{\{source\}} is the original paragraph for rewriting and the rendered evidence sheet for writing from evidence.

Label $\lab{1}$ depends on what a sentence does, not on whether the source already says it, so a faithful paraphrase of the authors' own not-shown sentence is also $\lab{1}$. This affects few texts. For each $\lab{1}$ sentence the coder also quotes the span of the source that asserts what the sentence says is not shown, which would separate retracting a claim of the source from denying a conclusion the source never drew. The insertion test shows that it cannot make this separation reliably (Appendix~\ref{app:coder-val}), so the main text merges both into $\lab{1}$.

\begin{promptbox}[title=Sentence coder prompt]
Below is SOURCE material that a writer had, followed by a TEXT the writer produced from it, split into numbered sentences. The text comes from a computer science research paper.\\[1ex]
Give every sentence exactly one label. If a sentence does several things, use the first label in this list that applies to any part of it:\\
1. not\_shown: the sentence says what the evidence or results do not show, establish, imply, isolate, explain, or guarantee, or how they should not be interpreted ("alone does not establish", "should not be interpreted as", "rather than establishing general superiority", "does not by itself explain", "remains to be established"). It also covers directives on how the results should be read or interpreted ("should be interpreted as targeted stress tests", "should be read with the caveat that", "is best viewed as"). This is a statement about what can be concluded. A factual negative result is not not\_shown: "our method does not outperform CP on Wiki80" or "it falls short of Fish on Cartoon" reports a result and is grounded\_limit (or unsupported\_doubt if SOURCE does not contain it). For not\_shown sentences also give "proposition" (what is said not to be shown) and "source\_quote" (an exact span of SOURCE that asserts that proposition, or "" if SOURCE does not assert it).\\
2. unsupported\_doubt: the sentence names a limitation, confound, alternative explanation, or needed validation that SOURCE does not contain ("although training costs may differ", "confirming this would require controlled ablations").\\
3. grounded\_limit: the sentence states, anywhere in it, a limitation, cost, requirement, condition, or a mixed or unfavorable result (the method trails, ties, or only marginally beats a baseline somewhere) that SOURCE contains. A sentence that mostly reports a win but adds "although Fish is best on average" is grounded\_limit.\\
4. generic\_scope: the sentence's only qualification narrows a claim to the evaluated, tested, or reported settings without naming them ("in the evaluated settings, X improves Y", "within the scope of these experiments"). A phrase asserting breadth, such as "consistently outperforms across all evaluated settings", is a claim, not generic\_scope. A sentence that names specific datasets, settings, or conditions is not generic\_scope.\\
5. future\_work: the sentence proposes future work or open directions.\\
6. claim: the sentence reports a favorable result or states a claim, interpretation, or contribution, with none of the above.\\
7. other: background, method or setup description, transitions.\\[1ex]
SOURCE:\\
\{source\}\\[1ex]
TEXT:\\
\{sentences\}\\[1ex]
Return only a JSON object with one entry per sentence, in order:\\
\{"sentences": [\{"id": 1, "label": "<label>", "proposition": "", "source\_quote": ""\}]\}
\end{promptbox}

\subsection{Coder Validation}
\label{app:coder-val}

\paragraph{Insertion test.}
We plant sentences of known type into the authors' paragraphs and check whether the coder recovers the type. Kimi-K2.6, which is neither the coder nor one of the writers, receives each of the 77 source paragraphs with its list of claims (Appendix~\ref{app:claims}) and writes six sentences that fit the paragraph: a \emph{retraction} saying that one of the listed claims is not shown, a \emph{denial} saying that the results do not show a stronger conclusion the paragraph never draws, an ungrounded doubt, a generic scope, a grounded limitation, and a claim. The retraction and denial use phrasings such as ``does not by itself establish'' and ``should not be interpreted as''. Paragraphs that state no limitation get no grounded limitation, which leaves 441 sentences. Each sentence is inserted at a random sentence boundary of its paragraph, and the coder labels the modified paragraph with the unmodified one as source, as in rewriting.

The coder labels 97\% (300/308) of the planted retractions, denials, ungrounded doubts, and generic scopes as $U$, and only 6\% (8/133) of the planted claims and grounded limitations. It labels 96\% (148/154) of the retractions and denials as $\lab{1}$, but its quote-based split assigns only 31\% of them to the right one of the two, so the main text does not separate them. Inserting a sentence rarely changes the labels of the others: 93\% (2{,}425/2{,}604) of the authors' sentences keep the label they receive in the unmodified paragraph.

\paragraph{Second coder.}
Qwen3-32B, run locally, recodes all rewrites with the same prompt and a system prompt that restates the label rules as checks. On rewriting under the default prompt, the two coders agree on whether a sentence is in $U$ for 97.2\% of sentences ($\kappa=0.66$). Their values of $R_U$ for the 12 writers and the authors' originals have a Spearman correlation of 0.80, and GPT-6-astra has the highest $R_U$ under both. On the prompt-context rewrites of \S\ref{sec:rq2}, $\kappa=0.63$.

\subsection{Claim Tracking}
\label{app:claims}

Claim tracking has two steps, both run by DeepSeek-V4-Flash at temperature 0. The coder first lists the claims in each original paragraph, once per paragraph. It then receives the original, the numbered claims, and one rewrite, and gives each claim a status. It does not know which model or prompt produced the rewrite. When several statuses apply, it chooses the one that changes the claim most.

\begin{promptbox}[title=Claim extraction prompt]
Below is a paragraph from a computer science research paper. List the claims the authors make in it: every statement about their method, their results, explanations of results, prior work, or the problem, that a reader could agree or disagree with. Keep each claim close to the authors' wording and keep its original strength. Do not list pure descriptions of procedure or setup (for example "we train for 10 epochs").\\[1ex]
PARAGRAPH:\\
\{text\}\\[1ex]
Return only a JSON object: \{"claims": ["<claim 1>", "<claim 2>"]\}
\end{promptbox}

\begin{promptbox}[title=Claim tracking prompt]
Below is a paragraph from a computer science research paper (ORIGINAL), a numbered list of the claims the authors make in it, and a rewrite of the paragraph (REVISED). For each claim, say what REVISED does with it:\\
- kept: REVISED states the claim with about the same strength and scope (wording may differ).\\
- strengthened: REVISED states it more strongly, more broadly, or more definitely.\\
- weakened: REVISED still asserts the claim but makes it tentative, partial, or softer ("causes" -> "may contribute to", "fails" -> "may struggle").\\
- scoped: REVISED keeps the claim but confines it to the tested settings, data, or conditions where the original did not.\\
- retracted: REVISED says the claim is not shown, not established, does not hold, or holds only in a narrower form than the authors stated, or argues against it.\\
- dropped: REVISED no longer contains the claim.\\
If several apply, choose the one that changes the claim most (retracted > dropped > weakened > scoped > strengthened > kept).\\[1ex]
ORIGINAL:\\
\{original\}\\[1ex]
CLAIMS:\\
\{claims\}\\[1ex]
REVISED:\\
\{revised\}\\[1ex]
Return only a JSON object: \{"claims": [\{"id": <claim number>, "status": "<kept|\allowbreak{}strengthened|\allowbreak{}weakened|\allowbreak{}scoped|\allowbreak{}retracted|\allowbreak{}dropped>", "revised\_span": "<shortest exact span from REVISED that expresses the claim or its change, or empty if dropped>"\}]\}
\end{promptbox}

\paragraph{Agreement.}
We sample five paragraphs of each type and take their rewrites under the default prompt by six writers: GPT-4o, GPT-5.2, GPT-5.5, GPT-5.6-sol, GPT-6-astra, and Claude Sonnet~5. This gives 120 rewrites with 762 claims. Three coders track them again with the same claim lists: DeepSeek-V4-Flash in a second run, Qwen3-32B, and Kimi-K2.6. Table~\ref{tab:track-agree} compares each with the main run.

\begin{table}[h]
\centering
\small
\setlength{\tabcolsep}{4pt}
\begin{tabular}{@{}lccccc@{}}
\toprule
& \multicolumn{2}{c}{Six statuses} & Soft & Ret & \\
\cmidrule(lr){2-3}
Coder & Agree & $\kappa$ & $\kappa$ & $\kappa$ & \#Ret \\
\midrule
DeepSeek-V4-Flash & 0.87 & 0.72 & 0.74 & 0.87 & 12 \\
Qwen3-32B & 0.73 & 0.49 & 0.59 & 0.35 & 17 \\
Kimi-K2.6 & 0.81 & 0.62 & 0.63 & 0.71 & 17 \\
\bottomrule
\end{tabular}
\caption{Agreement of claim tracking with the main run on 762 claims. Soft and Ret are binary: softened (weakened or scoped) or not, retracted or not. \#Ret is the number of retracted claims; the main run has 11.}
\label{tab:track-agree}
\end{table}

The second run of the main coder is close to the first. Under all three coders, GPT-6-astra has the most retracted claims, and it also has the most softened claims, though under Kimi-K2.6 only by a small margin. Both other coders find more retractions than the main coder. Most disagreements fall between \emph{weakened} and \emph{retracted}, on sentences such as ``does not by itself guarantee the removal of all class information'', which the main coder labels weakened.

\subsection{Claim-Support Judge}
\label{app:judge}

The judge, DeepSeek-V4-Flash at temperature 0, gives each claim extracted in Appendix~\ref{app:claims} one of five labels: \emph{supported} by the reported results at the stated strength and scope; \emph{overstated}, supported only in a weaker or narrower form; \emph{contradicted}; \emph{interpretation}, explaining why a result occurs, which the results neither confirm nor refute; and \emph{not covered} by the sheet. Only results and facts count as evidence. The judge sees the evidence sheet, the original paragraph, and the claims, but no rewrite.

\paragraph{Human check.} Two PhD students labeled 40 claims sampled by the judge's label: 12 supported, 12 overstated or contradicted, 8 interpretation, and 8 not covered. GPT retracted 39 of them in at least one setting of Table~\ref{tab:rq3}. The students saw the same input as the judge, did not see its labels or any rewrite, and labeled independently. They agree with each other on 30 of the 40 claims ($\kappa=0.66$), and with the judge on 19 and 21. By the judge's label, both students agree with it on 9 of 12 supported claims, 4 of 8 interpretations, 4 of 8 not-covered claims, and none of the 12 overstated or contradicted claims. Against the evidence sheet, the judge's supported labels hold up, but its overstated labels do not (for a check against the full papers, see Appendix~\ref{app:agent}), so we do not rely on the comparison between overstated and supported claims. Because the sample oversamples rare labels, these figures do not estimate agreement over all claims.

\begin{promptbox}[title=Claim-support judge prompt]
Below is an evidence sheet for a computer science paper, built from its method, experiment and related-work sections. It records the method, prior work, setup, every reported result with exact numbers, analysis outcomes, and the conditions the authors state. Then comes a paragraph from the same paper and a numbered list of claims the authors make in it.\\[1ex]
For each claim, decide what the evidence sheet says about it:\\
- supported: the reported results, numbers, analysis outcomes, or described facts of the method and prior work establish the claim at the strength and scope the authors state.\\
- overstated: the results support only a weaker or narrower version. Examples: the claim generalizes beyond the settings or cases that were tested; it says "consistently", "significantly", "exactly", or "always" but the results tie, lose, vary, or include no test; it states a cause where the results only show an association.\\
- contradicted: the results show the claim is false.\\
- interpretation: the claim explains why a result happens, or judges its meaning or importance, and the reported results neither confirm nor refute it.\\
- not\_covered: the claim concerns facts the sheet does not contain.\\
Only results and facts count as evidence. Some sheet entries, especially stated conditions, notes, and prior-work descriptions, repeat the authors' own judgment or wording; such a repetition is not evidence for the same judgment. A claim that judges prior work (for example that a method fails or does not scale) needs a measured result. Judge each claim only against the sheet. The paragraph is there to make clear what each claim refers to; it is not evidence.\\[1ex]
EVIDENCE SHEET:\\
\{sheet\}\\[1ex]
PARAGRAPH:\\
\{paragraph\}\\[1ex]
CLAIMS:\\
\{claims\}\\[1ex]
Return only a JSON object: \{"claims": [\{"id": <claim number>, "label": "<supported|\allowbreak{}overstated|\allowbreak{}contradicted|\allowbreak{}interpretation|\allowbreak{}not\_covered>", "evidence": "<the sheet item(s) your label rests on, quoted briefly, or empty>"\}]\}
\end{promptbox}

\section{Additional Results}
\label{app:results}

\subsection{Ungrounded Qualifications by Label}
\label{app:labels}

Table~\ref{tab:labels} splits $R_U$ in rewriting into its three labels. The rise across GPT versions comes mostly from not-shown statements. Generic scope, which can be a fair statement of the tested range, rises much less.

\begin{table}[h]
\centering
\small
\begin{tabular}{@{}lccc@{}}
\toprule
Writer & $R_{\lab{1}}$ & $R_{\lab{2}}$ & $R_{\lab{4}}$ \\
\midrule
Authors (original) & 0.03 & 0.00 & 0.01 \\
\midrule
GPT-4o & 0.00 & 0.00 & 0.02 \\
GPT-4.1 & 0.03 & 0.02 & 0.04 \\
GPT-5.1 & 0.04 & 0.01 & 0.03 \\
GPT-5.2 & 0.04 & 0.03 & 0.04 \\
GPT-5.4 & 0.09 & 0.01 & 0.04 \\
GPT-5.5 & 0.17 & 0.06 & 0.06 \\
GPT-5.6-sol & 0.09 & 0.10 & 0.12 \\
GPT-6-astra & 0.66 & 0.14 & 0.13 \\
\midrule
Claude Sonnet 5 & 0.04 & 0.05 & 0.01 \\
Claude Opus 5.5 & 0.04 & 0.01 & 0.00 \\
DeepSeek-V4-Pro & 0.03 & 0.02 & 0.00 \\
Grok 4.7 & 0.15 & 0.04 & 0.05 \\
\bottomrule
\end{tabular}
\caption{Not-shown statements ($\lab{1}$), ungrounded doubt ($\lab{2}$), and generic scope ($\lab{4}$) per 100 words in rewriting under the default prompt, pooled over the four paragraph types.}
\label{tab:labels}
\end{table}

\subsection{Prompt Contexts}
\label{app:context}

Table~\ref{tab:minimal} compares the polish-only prompt with the default prompt of \S\ref{sec:rq1} for all 12 models. Table~\ref{tab:gradient-claims} gives the claim-tracking results for the seven prompt contexts.

\paragraph{Revision without review.} We ran P8 and P9 after the other contexts, with the same models and high reasoning effort, and reran P4 for GPT-6-astra and GPT-5.5 at the same time. The rerun gives $R_U$ of 0.77 for GPT-6-astra and 0.20 for GPT-5.5, against 0.72 and 0.25 in the original run. Under P8, $R_U$ is 0.61 for GPT-6-astra, 0.23 for GPT-5.5, and 0.11 for Claude Opus~5.5 (P4: 0.77, 0.20, 0.11); under P9 it is 0.39, 0.24, and 0.05. GPT-6-astra retracts 6.2\%, 6.6\%, and 3.1\% of the authors' claims under P4, P8, and P9, and softens 33\%, 31\%, and 31\%. The judge (Appendix~\ref{app:judge}) labels 23\%, 22\%, and 27\% of the claims it retracts as supported.

\begin{table}[h]
\centering
\small
\begin{tabular}{lccc}
\toprule
& \multicolumn{2}{c}{Polish only} & Default \\
\cmidrule(lr){2-3}\cmidrule(lr){4-4}
Writer & $R_U$ & $P_U$\,\% & $R_U$ \\
\midrule
Authors (original) & 0.04 & 5 & -- \\
\midrule
GPT-4o & 0.03 & 4 & 0.02 \\
GPT-4.1 & 0.02 & 3 & 0.08 \\
GPT-5.1 & 0.04 & 5 & 0.07 \\
GPT-5.2 & 0.05 & 6 & 0.10 \\
GPT-5.4 & 0.03 & 4 & 0.14 \\
GPT-5.5 & 0.04 & 5 & 0.29 \\
GPT-5.6-sol & 0.02 & 3 & 0.31 \\
GPT-6-astra & 0.03 & 4 & 0.94 \\
Claude Sonnet 5 & 0.04 & 5 & 0.10 \\
Claude Opus 5.5 & 0.03 & 4 & 0.05 \\
DeepSeek-V4-Pro & 0.06 & 6 & 0.05 \\
Grok 4.7 & 0.04 & 5 & 0.23 \\
\bottomrule
\end{tabular}
\caption{Ungrounded qualifications when the 12 models are asked only to polish the paragraph, against the default prompt of \S\ref{sec:rq1}. All four paragraph types are pooled.}
\label{tab:minimal}
\end{table}

\begin{table*}[h]
\centering
\small
\begin{tabular}{lccccccc}
\toprule
Writer & P1 & P2 & P3 & P4 & P5 & P6 & P7 \\
\midrule
GPT-4o & 11 / 0.0 & 8 / 0.0 & 10 / 0.0 & 18 / 0.2 & 18 / 0.6 & 11 / 0.0 & 17 / 0.2 \\
GPT-5.2 & 10 / 0.0 & 11 / 0.0 & 10 / 0.0 & 22 / 0.2 & 26 / 0.4 & 15 / 0.2 & 26 / 0.4 \\
GPT-5.5 & 9 / 0.0 & 10 / 0.0 & 11 / 0.2 & 29 / 0.2 & 30 / 0.8 & 15 / 0.2 & 27 / 0.0 \\
GPT-5.6-sol & 8 / 0.0 & 9 / 0.2 & 9 / 0.2 & 25 / 0.4 & 30 / 2.7 & 16 / 0.0 & 26 / 1.0 \\
GPT-6-astra & 9 / 0.2 & 7 / 0.2 & 11 / 0.0 & 33 / 6.0 & 33 / 9.5 & 20 / 0.8 & 33 / 5.8 \\
Claude Opus 5.5 & 3 / 0.2 & 5 / 0.2 & 7 / 0.2 & 15 / 0.2 & 15 / 1.1 & 12 / 0.2 & 14 / 0.6 \\
\bottomrule
\end{tabular}
\caption{Share of the authors' claims softened / retracted (\%) under the seven prompt contexts of \S\ref{sec:rq2}.}
\label{tab:gradient-claims}
\end{table*}

\subsection{Multi-round Revision}
\label{app:multiround}

Table~\ref{tab:multiround-extra} gives paragraph length, the number of ungrounded qualifications per paragraph, softening, and dropped claims for the chains of Table~\ref{tab:multiround}. A claim is dropped when it no longer appears in the revision. We read 12 dropped claims from rounds 2 and 3. About a third are defensive: the claim is replaced by a statement that the evidence does not establish it, or by a disclaimer of novelty. About half disappear because the paragraph is rewritten into definitions and scope statements in response to review comments asking for precision. One or two are coder errors in which the claim survives in other words. About half of the claims dropped in rounds 2 and 3 were already dropped in the previous round. We therefore do not count dropped claims as defense. In later rounds, Claude also adds content that is not in the original, such as citations to architectures published after the paper.

\begin{table*}[h]
\centering
\small
\setlength{\tabcolsep}{3.5pt}
\begin{tabular}{ll ccc ccc ccc ccc}
\toprule
& & \multicolumn{3}{c}{Words} & \multicolumn{3}{c}{$U$ per paragraph} & \multicolumn{3}{c}{Soft\,\%} & \multicolumn{3}{c}{Dropped\,\%} \\
\cmidrule(lr){3-5}\cmidrule(lr){6-8}\cmidrule(lr){9-11}\cmidrule(lr){12-14}
Chain & Writer & R1 & R2 & R3 & R1 & R2 & R3 & R1 & R2 & R3 & R1 & R2 & R3 \\
\midrule
Review--revise & GPT-5.5 & 232 & 319 & 345 & 2.2 & 3.9 & 3.8 & 32 & 34 & 28 & 7 & 19 & 27 \\
Review--revise & GPT-5.6-sol & 234 & 338 & 363 & 3.3 & 5.5 & 5.2 & 32 & 28 & 24 & 9 & 20 & 30 \\
Review--revise & GPT-6-astra & 248 & 294 & 312 & 4.6 & 4.7 & 5.2 & 27 & 26 & 25 & 8 & 19 & 30 \\
Review--revise & Claude Opus 5.5 & 194 & 259 & 314 & 1.0 & 1.6 & 2.2 & 20 & 25 & 24 & 9 & 18 & 24 \\
\midrule
Polish only & GPT-5.6-sol & 119 & 117 & 117 & 0.1 & 0.1 & 0.0 & 10 & 11 & 11 & 0 & 0 & 0 \\
Polish only & GPT-6-astra & 117 & 115 & 114 & 0.1 & 0.0 & 0.0 & 6 & 8 & 7 & 0 & 1 & 0 \\
Polish only & Claude Opus 5.5 & 117 & 114 & 112 & 0.0 & 0.0 & 0.1 & 5 & 6 & 8 & 1 & 1 & 2 \\
\bottomrule
\end{tabular}
\caption{Further measures for the multi-round chains of Table~\ref{tab:multiround}. $U$ per paragraph is the average number of ungrounded qualifications in a paragraph. The authors' originals average 128 words.}
\label{tab:multiround-extra}
\end{table*}

\subsection{Claim Support}
\label{app:support}

The judge labels all 482 claims of the 77 original paragraphs: 52\% supported, 5\% overstated or contradicted, 15\% interpretation, and 28\% not covered. Let $C_L$ be the claims with label $L$. Table~\ref{tab:rq3} gives $\mathrm{Ret}_L(w)$ and $\mathrm{Soft}_L(w)$, Eqs.~\eqref{eq:ret} and~\eqref{eq:soft} computed over $C_L$ instead of $C$. Because few claims are retracted under the default prompt, it also covers the first round of review--revise chains (\S\ref{sec:rq2}).

By these labels, GPT softens overstated claims two to two and a half times as often as supported ones and retracts them at least twice as often, but about a third of GPT-6-astra's retractions under the default prompt and under P5, and 37\% of the GPT versions' retractions in the first review round, fall on claims labeled supported. Annotators do not confirm the overstated labels (Appendix~\ref{app:judge}), so these comparisons are not evidence either way. Table~\ref{tab:rq3} uses only the paragraphs for which the claim tracker and the judge list the same claims, 71 of 77 under the default prompt. For this reason GPT-6-astra has 34 retracted claims here and 38 in Table~\ref{tab:rq1}.

\begin{table*}[t]
\centering
\small
\setlength{\tabcolsep}{4pt}
\begin{tabular}{@{}l cc ccc ccc@{}}
\toprule
& \multicolumn{2}{c}{Soft\,\%} & \multicolumn{3}{c}{Ret\,\%} & \multicolumn{2}{c}{Retracted claims\,\%} & \\
\cmidrule(lr){2-3}\cmidrule(lr){4-6}\cmidrule(lr){7-8}
Writer & Sup & Over & Sup & Over & Int & Sup & Over & $n$ \\
\midrule
GPT-5.2 to 6-astra & 21 {\scriptsize[18, 24]} & 53 {\scriptsize[42, 65]} & 1.2 {\scriptsize[0.5, 2.0]} & 5.7 {\scriptsize[0.0, 17.5]} & 2.4 {\scriptsize[0.7, 4.7]} & 30 {\scriptsize[16, 46]} & 13 {\scriptsize[0, 29]} & 46 \\
GPT-6-astra & 35 {\scriptsize[28, 42]} & 71 {\scriptsize[59, 88]} & 4.7 {\scriptsize[1.9, 8.0]} & 9.5 {\scriptsize[0.0, 23.8]} & 10.6 {\scriptsize[2.9, 19.2]} & 32 {\scriptsize[17, 49]} & 6 {\scriptsize[0, 13]} & 34 \\
GPT-6-astra, P5 & 27 {\scriptsize[19, 36]} & 68 {\scriptsize[55, 81]} & 5.6 {\scriptsize[3.1, 8.3]} & 13.6 {\scriptsize[0.0, 31.2]} & 20.5 {\scriptsize[10.2, 31.2]} & 30 {\scriptsize[20, 43]} & 7 {\scriptsize[0, 13]} & 46 \\
GPT, R1 & 32 {\scriptsize[26, 38]} & 48 {\scriptsize[33, 67]} & 13.0 {\scriptsize[9.8, 16.3]} & 25.8 {\scriptsize[12.7, 43.9]} & 33.3 {\scriptsize[23.9, 42.6]} & 37 {\scriptsize[30, 46]} & 6 {\scriptsize[3, 11]} & 262 \\
Claude Opus, R1 & 18 {\scriptsize[13, 23]} & 36 {\scriptsize[20, 64]} & 0.0 {\scriptsize[0.0, 0.0]} & 4.5 {\scriptsize[0.0, 15.8]} & 4.1 {\scriptsize[0.0, 9.1]} & -- & -- & 6 \\
Controls & 9 {\scriptsize[7, 12]} & 31 {\scriptsize[20, 43]} & 0.3 {\scriptsize[0.0, 0.7]} & 4.8 {\scriptsize[0.0, 13.3]} & 0.4 {\scriptsize[0.0, 1.3]} & 27 {\scriptsize[0, 64]} & 36 {\scriptsize[0, 75]} & 11 \\
\bottomrule
\end{tabular}
\caption{What rewrites do to the authors' claims, by the judge's label: Sup = supported, Over = overstated or contradicted, Int = interpretation. Soft and Ret: share of the claims with each label that are softened or retracted. Retracted claims: share of the retracted claims that carry each label; $n$ is the number of retracted claims, and shares are omitted when $n<10$. R1: first round of review--revise chains, with GPT pooling GPT-5.5, GPT-5.6-sol, and GPT-6-astra. Controls: the four non-GPT models. Annotators confirm the judge's supported labels but not its overstated labels. Brackets are 95\% bootstrap intervals over papers.}
\label{tab:rq3}
\end{table*}

\subsection{Human Check against the Full Papers}
\label{app:agent}

\paragraph{Materials.} The check covers GPT-6-astra's P5 rewrites. It covers 16 papers and includes all 46 claims that claim tracking marks as retracted and 12 claims from the same paragraphs that GPT-6-astra keeps or softens, so that the target claims do not reveal which were retracted. It also includes 60 ungrounded qualifications from P5 rewrites: 36 by GPT-6-astra, 16 not-shown statements, 10 ungrounded doubts and 10 generic-scope sentences, and 12 each by GPT-5.5 and Claude Opus~5.5. Each paper comes with its parsed full text and tables and a link to its PDF. Rewrites are labeled only by number, and the writer, prompt, coder labels, and tracking status are withheld.

\paragraph{Procedure.} Two PhD students, other than those in Appendix~\ref{app:judge}, each work through the whole packet independently, using a web annotation form and our annotation guide. For each claim they judge whether the paper supports it (supported, overstated, contradicted, interpretation, cannot tell); for each rewrite, what it does to the claim (the six tracking statuses) and whether that change is warranted, overcorrected, unnecessary, or wrong. For each qualification they judge whether the paper raises its point and whether it is warranted, redundant, or wrong. Redundant means that the sentence is not wrong but tells a careful reader nothing new. Each answer comes with a short reason. We read a sample of the reasons.

\paragraph{Results.} Table~\ref{tab:agent} gives agreement and the counts on which both annotators agree. Of the 58 rewrites, both call 18 retractions and at least one calls 28, whereas claim tracking marks 46 of them retracted.

\begin{table}[t]
\centering
\footnotesize
\setlength{\tabcolsep}{3pt}
\begin{tabular}{@{}lcc@{}}
\toprule
Question & Agreement & $\kappa$ \\
\midrule
Claim supported by the paper & 0.84 & 0.78 \\
What the rewrite does & 0.78 & 0.69 \\
Change warranted & 0.83 & 0.71 \\
Qualification raised by the paper & 0.80 & 0.68 \\
Qualification warranted & 0.92 & 0.83 \\
\midrule
\multicolumn{3}{@{}l}{\emph{Claims both find changed (48), both agree}} \\
Supported / overstated$^{*}$ / interpretation & \multicolumn{2}{c}{4 / 20 / 15} \\
Change warranted / unnecessary & \multicolumn{2}{c}{28 / 10} \\
\multicolumn{3}{@{}l}{\emph{Claims both find retracted (18), both agree}} \\
Supported / overstated$^{*}$ / interpretation & \multicolumn{2}{c}{0 / 8 / 8} \\
Retraction warranted / unnecessary & \multicolumn{2}{c}{13 / 1} \\
\multicolumn{3}{@{}l}{\emph{Qualifications (60), both agree}} \\
Redundant / warranted / wrong & \multicolumn{2}{c}{36 / 16 / 3} \\
\bottomrule
\end{tabular}
\caption{Check of P5 rewrites against the full papers by two PhD annotators. $^{*}$Overstated or contradicted.}
\label{tab:agent}
\end{table}

\section{Reviewers and Readers}
\label{app:rq4}

\subsection{AI Reviewer Probes}
\label{app:probes}

Claude Sonnet~5 and DeepSeek-V4-Flash review each version independently with their default settings and do not know where it comes from. In the prompt below, \texttt{\{where\}} gives the position of the paragraph in the paper according to its type. The comparison uses the 73 paragraphs for which all compared rewrites were available when we ran the reviews. For each writer we keep the paragraphs whose P1 and P5 rewrites have valid reviews from both models, which leaves 68 to 72.

Table~\ref{tab:probe-scores} gives the mean scores from each reviewer model. Under both models, the P5 rewrites of all four writers score higher than their P1 rewrites on all four aspects.

DeepSeek-V4-Flash sorts the listed weaknesses into seven categories, seeing only the text of the weaknesses and not the paragraph, the writer, or the prompt. Table~\ref{tab:weak-cats} gives the number of weaknesses per review in each category. P5 rewrites receive slightly fewer weaknesses, mostly because they are criticized less for overclaiming. The P5 rewrites of GPT-6-astra are criticized more often for two things: being overly cautious, and weak significance. Some of the latter say that the paragraph disclaims novelty without saying what is new. The reviewers also criticize these rewrites less often for being unclear and give them higher clarity scores, while human readers find them harder to read (\S\ref{sec:rq4}).

Among reviews that criticize a rewrite as overly cautious, the P5 rewrites of the three GPT versions still receive higher overall scores on average than their P1 rewrites, 6.50 to 6.79 against 5.72 to 5.81.

\begin{promptbox}[title=Reviewer prompt]
You are a reviewer for a top-tier machine learning / NLP conference. Below is one paragraph from \{where\} of a submission. Assume that the numbers it reports are correct and that the rest of the paper is competent; judge this paragraph as written.\\[1ex]
PARAGRAPH:\\
\{text\}\\[1ex]
Rate the paragraph from 1 (very poor) to 10 (excellent) on:\\
- soundness: are the claims supported by and appropriately matched to the evidence the paragraph presents?\\
- contribution: how significant and clearly conveyed is the finding or contribution?\\
- clarity: how easily can a reader tell what the authors found and what they claim?\\
- overall: your overall assessment of this paragraph for a top-tier paper.\\
Then list the main weaknesses a reviewer would raise (at most three, empty if none).\\[1ex]
Return only a JSON object:\\
\{"soundness": <1-10>, "contribution": <1-10>, "clarity": <1-10>, "overall": <1-10>, "weaknesses": ["..."]\}
\end{promptbox}

\begin{promptbox}[title=Weakness categorization prompt]
Below are the weaknesses a peer reviewer listed for one paragraph of a computer science paper. Put each weakness into exactly one category:\\
- overclaim: a claim is stronger than the evidence shown, an explanation or causal claim is unsupported or speculative, or a result is generalized too far.\\
- missing\_detail: numbers, statistics, significance tests, variance, baselines, setup details, or definitions are missing.\\
- unclear: wording, terms, or structure are vague, ambiguous, or confusing.\\
- shallow: the analysis is superficial, does not explain why a result occurs, or proposes no follow-up.\\
- significance: the novelty, contribution, motivation, or positioning against prior work is weak.\\
- overhedged: the paragraph is too cautious, has too many caveats or disclaimers, or buries its contribution.\\
- other: none of the above.\\[1ex]
WEAKNESSES:\\
\{items\}\\[1ex]
Return only a JSON object: \{"categories": ["<category of weakness 1>", "<category of weakness 2>", ...]\}
\end{promptbox}

\begin{table}[t]
\centering
\footnotesize
\setlength{\tabcolsep}{1.6pt}
\begin{tabular}{@{}lcccccccc@{}}
\toprule
& \multicolumn{2}{c}{GPT-5.5} & \multicolumn{2}{c}{GPT-5.6-sol} & \multicolumn{2}{c}{GPT-6-astra} & \multicolumn{2}{c}{Opus 5.5} \\
\cmidrule(lr){2-3}\cmidrule(lr){4-5}\cmidrule(lr){6-7}\cmidrule(lr){8-9}
& P1 & P5 & P1 & P5 & P1 & P5 & P1 & P5 \\
\midrule
\multicolumn{9}{@{}l}{\emph{Claude Sonnet~5}} \\
Soundness & 4.91 & 6.71 & 5.01 & 6.35 & 5.04 & 7.49 & 4.96 & 5.72 \\
Contribution & 4.61 & 5.66 & 4.52 & 5.25 & 4.61 & 5.62 & 4.59 & 5.07 \\
Clarity & 5.10 & 6.74 & 5.32 & 6.35 & 5.35 & 6.96 & 5.29 & 6.24 \\
Overall & 4.77 & 6.20 & 4.87 & 5.89 & 4.86 & 6.47 & 4.85 & 5.56 \\
\addlinespace
\multicolumn{9}{@{}l}{\emph{DeepSeek-V4-Flash}} \\
Soundness & 6.53 & 7.70 & 6.54 & 7.61 & 6.71 & 7.94 & 6.69 & 7.26 \\
Contribution & 6.34 & 7.34 & 6.28 & 7.01 & 6.25 & 6.78 & 6.49 & 7.04 \\
Clarity & 7.50 & 8.57 & 7.45 & 8.28 & 7.64 & 8.47 & 7.49 & 8.25 \\
Overall & 6.67 & 7.81 & 6.62 & 7.59 & 6.76 & 7.58 & 6.72 & 7.40 \\
\midrule
Paragraphs & \multicolumn{2}{c}{70} & \multicolumn{2}{c}{71} & \multicolumn{2}{c}{72} & \multicolumn{2}{c}{68} \\
\bottomrule
\end{tabular}
\caption{Mean scores (1--10) from each reviewer model for P1 and P5 rewrites.}
\label{tab:probe-scores}
\end{table}

\begin{table}[t]
\centering
\footnotesize
\setlength{\tabcolsep}{1.4pt}
\begin{tabular}{@{}lcccccccc@{}}
\toprule
& \multicolumn{2}{c}{GPT-5.5} & \multicolumn{2}{c}{GPT-5.6-sol} & \multicolumn{2}{c}{GPT-6-astra} & \multicolumn{2}{c}{Opus 5.5} \\
\cmidrule(lr){2-3}\cmidrule(lr){4-5}\cmidrule(lr){6-7}\cmidrule(lr){8-9}
& P1 & P5 & P1 & P5 & P1 & P5 & P1 & P5 \\
\midrule
Overclaim & 0.79 & 0.56 & 0.78 & 0.59 & 0.73 & 0.31 & 0.79 & 0.77 \\
Missing detail & 0.89 & 0.91 & 0.91 & 0.91 & 0.86 & 0.85 & 0.92 & 0.93 \\
Unclear & 0.61 & 0.52 & 0.58 & 0.54 & 0.59 & 0.42 & 0.51 & 0.46 \\
Shallow & 0.19 & 0.16 & 0.15 & 0.19 & 0.20 & 0.20 & 0.20 & 0.21 \\
Significance & 0.41 & 0.37 & 0.45 & 0.43 & 0.44 & 0.60 & 0.43 & 0.36 \\
Overhedged & 0.04 & 0.09 & 0.02 & 0.08 & 0.05 & 0.26 & 0.04 & 0.07 \\
Other & 0.01 & 0.00 & 0.02 & 0.00 & 0.01 & 0.00 & 0.01 & 0.01 \\
\midrule
Total & 2.94 & 2.60 & 2.91 & 2.74 & 2.88 & 2.65 & 2.90 & 2.80 \\
\bottomrule
\end{tabular}
\caption{Weaknesses per review in each category, averaged over the two reviewer models.}
\label{tab:weak-cats}
\end{table}

\subsection{Minimal Pairs}
\label{app:minpair}

We trim every P5 rewrite that has at least one sentence in $U$ and a valid review in the comparison above. Grok~4.7 receives the rewrite and the list of its sentences in $U$, with the prompt below. Of 147 rewrites, 3 failed because the API timed out. In 143 of the other 144, every word of the trimmed version appears in the P5 rewrite; in the last one, the editor moved part of a deleted sentence into the next sentence. Deleting whole sentences instead would also remove the results that some of these sentences report, and sometimes leaves a pronoun without a referent. The coder then labels the trimmed texts in the same way as all other texts. Both reviewer models review each P5 rewrite and its trimmed version twice, with the reviewer prompt above; we average the two reviews and take paired differences, with bootstrap over papers.

Table~\ref{tab:minpair} gives the results by writer. The drop is larger under Claude Sonnet~5 than under DeepSeek-V4-Flash, but it is present for every writer under both. In the third of pairs in which trimming removes the fewest words, 2 to 15\%, overall drops by 0.41 [0.31, 0.50]; in the middle third, 15 to 23\%, by 0.82; and in the top third, 23 to 56\%, by 1.24. The P5 scores in this run match those of the main comparison: on the same paragraphs, the mean overall score of GPT-6-astra's P5 rewrites is 7.03, against 7.09 before.

\begin{table}[t]
\centering
\footnotesize
\setlength{\tabcolsep}{2.2pt}
\begin{tabular}{@{}lcccccc@{}}
\toprule
& & \multicolumn{2}{c}{$R_U$} & \multicolumn{3}{c}{Trimmed $-$ P5} \\
\cmidrule(lr){3-4}\cmidrule(l){5-7}
Writer & $n$ & P5 & Trim. & Overall & Sound. & Contr. \\
\midrule
GPT-5.5 & 35 & 0.95 & 0.07 & $-$0.68 & $-$0.88 & $-$0.61 \\
GPT-5.6-sol & 29 & 1.04 & 0.25 & $-$0.93 & $-$1.14 & $-$0.67 \\
GPT-6-astra & 57 & 1.41 & 0.18 & $-$0.95 & $-$1.34 & $-$0.65 \\
Opus 5.5 & 23 & 0.93 & 0.15 & $-$0.61 & $-$0.92 & $-$0.50 \\
\midrule
All & 144 & & & $-$0.82 & $-$1.12 & $-$0.62 \\
\bottomrule
\end{tabular}
\caption{AI review of P5 rewrites with only their ungrounded qualifications removed. Scores are from 1 to 10, averaged over the two reviewer models and two reviews each; $n$ is the number of pairs. Trimming shortens the rewrites by 17 to 23\% of their words. 95\% CIs for All: overall [$-$0.95, $-$0.70], soundness [$-$1.29, $-$0.97], contribution [$-$0.76, $-$0.50].}
\label{tab:minpair}
\end{table}

\begin{promptbox}[title=Trimming prompt]
Below is a paragraph from a research paper and a list of FLAGGED sentences from it. Each flagged sentence contains a qualification: a statement of what the results do not show or establish, a caveat or needed validation, or a phrase that narrows a claim to "the evaluated settings" or similar.\\[1ex]
Edit the paragraph as follows.\\
- In each flagged sentence, delete only the qualification. If the sentence also states a result, claim, or other content, keep that content with its original wording. Example: "X improves accuracy on both datasets, although this does not by itself establish that X generalizes." becomes "X improves accuracy on both datasets." Example: "In the evaluated settings, X outperforms Y." becomes "X outperforms Y."\\
- If a flagged sentence is nothing but the qualification, delete the whole sentence.\\
- Leave every unflagged sentence exactly as it is, except for the smallest grammatical repair that a deletion makes necessary, such as removing a connective like "also" or "however" that no longer connects to anything, or replacing a pronoun whose referent was deleted with the noun it referred to.\\
- Do not add any new words, claims, qualifications, or emphasis. Do not strengthen any claim beyond removing the qualification.\\[1ex]
PARAGRAPH:\\
\{text\}\\[1ex]
FLAGGED SENTENCES:\\
\{flagged\}\\[1ex]
Return only JSON: \{"text": "<edited paragraph>"\}
\end{promptbox}

\subsection{Human Reader Study}
\label{app:readers}

\paragraph{Materials and assignment.}
The 20 paragraphs are 10 result-analysis and 10 discussion paragraphs drawn at random from those with complete rewrites. Each appears in two versions, the P1 and P5 rewrites by GPT-6-astra, shown in full. Each of the four readers read all 20 paragraphs, ten in each version, and never saw both versions of the same paragraph. Each version of each paragraph was read by two readers, which gives 80 readings. Readers received anonymous texts and neutral task instructions. They were not told the version labels, the model that wrote the texts, the rewriting prompts, our hypotheses, or any earlier results.

\paragraph{Procedure.}
For each paragraph, a reader first read the whole paragraph and gave five ratings: how easy it is to read, how smoothly its main line of argument flows, how much effort it takes to understand, how easily its main claim can be identified, and how certain the authors are about that claim. The first four are rated from 1 to 7 and author certainty from 1 to 5. Higher effort means more effort; on the other scales, higher values mean more of the property. The reader then restated the main claim in their own words and answered two comprehension questions about information that both versions contain. Each step was locked once submitted, so later steps could not change earlier answers.

\paragraph{Analysis.}
For each paragraph and version we average the ratings of the two readers, which gives 20 paragraph-level differences between P5 and P1 for each rating. The intervals in Table~\ref{tab:rq4-human} are percentile intervals from 2{,}000 bootstrap resamples of these differences with seed 0. They describe variation across paragraphs for these four readers, not across readers. The restatements were collected but not scored, so claim identifiability is the readers' own rating and does not show whether they recovered the claim. Readers answered every comprehension question correctly for both versions, so this measure is at ceiling and cannot detect a difference.

\section{Examples}
\label{app:examples}

Figure~\ref{fig:examples} gives one GPT-6-astra rewrite under the default prompt for each paragraph type. In panel~(a), the authors' original contains no ungrounded qualification. GPT-5.5 adds only ``across the evaluated configurations'' to the summary sentence. GPT-6-astra adds two generic-scope sentences and two not-shown statements, and retracts the authors' explanation of the short-sentence result; checked against the full paper, the same retraction in its P5 rewrite is warranted (\S\ref{sec:rq3}). In the novelty paragraph, GPT-6-astra drops the 30--50\% figure and adds four not-shown statements; one of them turns the authors' conclusion about heuristic datasets into a conclusion that should not be drawn. In the contribution paragraph, it drops the state-of-the-art claim, limits the comparison with baselines to the evaluated settings, and adds that the method's threshold does not guarantee what it is designed to measure. In the discussion paragraph, it replaces the claim that ACML is a general model with a statement that the results do not establish general applicability.

\begin{figure*}[t]
\centering
\begin{minipage}[t]{0.49\textwidth}
\vspace{0pt}
\begin{examplebox}{(a) Result analysis \citep{ferrando-etal-2022-measuring}}
\who{Authors} It can be seen that across every different configuration, our proposed ALTI method outperforms other input attribution methods. [\dots] \emph{This is expected since these datasets are simpler, and therefore, interpretations can more easily find the important tokens.}
\tcblower
\who{GPT-5.5} \uline{Across the evaluated configurations}, ALTI achieves the strongest attribution performance among the compared methods.~$\lab{4}$ [\dots]
\tcbline
\who{GPT-6-astra} ALTI achieves the best scores on both metrics \uline{across the evaluated model--dataset configurations}.~$\lab{4}$ [\dots] \textbf{Although this pattern may be related to input length, these comparisons do not isolate its effect from other dataset characteristics and therefore do not establish that shorter inputs are inherently easier to interpret.}~$\lab{1}$ [\dots] \textbf{These results support ALTI's effectiveness under the evaluated token-removal metrics, rather than establishing the quality of its explanations more generally.}~$\lab{1}$
\end{examplebox}
\begin{examplebox}{(c) Contribution \citep{zhang-feng-2022-itst}}
\who{Authors} [\dots] Experiments on both text-to-text and speech-to-text simultaneous translation tasks show that ITST outperforms strong baselines and achieves state-of-the-art performance.
\tcblower
\who{GPT-6-astra} [\dots] \textbf{The threshold thus provides an explicit criterion for initiating translation, rather than a guarantee that sufficient information has been received.}~$\lab{1}$ Experiments on text-to-text and speech-to-text simultaneous translation show that ITST outperforms strong baselines \uline{in the evaluated settings}.~$\lab{4}$
\end{examplebox}
\end{minipage}\hfill
\begin{minipage}[t]{0.49\textwidth}
\vspace{0pt}
\begin{examplebox}{(b) Novelty \citep{wang-etal-2022-squality}}
\who{Authors} [\dots] However, studies have found that 30--50\% of summaries created this way contain facts that are unsupported by the rest of the article [CITE]. Models trained on this dataset learn to repeat this noise pattern by hallucinating facts in their outputs. \emph{It appears that known heuristics do not produce reliable data.}
\tcblower
\who{GPT-6-astra} [\dots] Prior analyses have identified reference summaries containing information that is not supported by the paired input [CITE]. \textbf{This information is not necessarily factually incorrect, but its absence from the input creates a mismatch for source-grounded summarization.}~$\lab{1}$ \textbf{Training on such pairs may encourage unsupported generation, although reference-level inconsistencies alone do not establish this causal effect.}~$\lab{1}$ \textbf{These observations motivate explicit checks of source--summary consistency during dataset construction, rather than a general conclusion that heuristic extraction methods cannot produce reliable data.}~$\lab{1}$
\end{examplebox}
\begin{examplebox}{(d) Discussion \citep{mao2020pruning}}
\who{Authors} [\dots] Both ACML-mean and ACML-attention improve ACML by a clear margin, and they perform much better than CommNet and AMP as shown in Table [REF]. \emph{The results imply that ACML is a general model}, and it can be easily extended with advanced message processing operations to achieve better results.
\tcblower
\who{GPT-6-astra} [\dots] As shown in Table [REF], both variants outperform the original ACML as well as CommNet and AMP. These results suggest that ACML can accommodate and benefit from alternative message-processing operations \uline{in the evaluated settings}.~$\lab{4}$ \textbf{Although these findings do not establish general applicability, they motivate further evaluation of ACML with additional message-processing operations and across a broader range of tasks.}~$\lab{1}$
\end{examplebox}
\end{minipage}
\caption{Rewrites under the default prompt, one paragraph of each type. [\dots]: omitted sentences. $\lab{4}$: generic scope, with the scoping phrase underlined. $\lab{1}$: not-shown statement, in bold. Italic: the authors' claim that the claim tracker labels retracted. [CITE] and [REF] are placeholders in the model input.}
\label{fig:examples}
\end{figure*}

\end{document}